\pdfoutput=1

\documentclass[11pt]{article}

\usepackage{acl}

\usepackage{float}
\usepackage{url}

\usepackage{times}
\usepackage{latexsym}
\usepackage{amssymb}
\usepackage{algorithm}
\usepackage{algorithmic}  
\usepackage[algo2e]{algorithm2e} 
\usepackage{amsmath}
\usepackage{multirow}
\usepackage{colortbl}
\usepackage{makecell}
\usepackage{comment}
\usepackage{graphicx}
\graphicspath{ {./images/} }

\newcommand{\NPTime}{{\sc NPTime}}
\newcommand{\NLogSpace}{{\sc NLogSpace}}
\newcommand{\ExpTime}{{\sc ExpTime}}
\newcommand{\NExpTime}{{\sc NExpTime}}

\newcommand{\Sat}{\ensuremath{\mathrm{Sat}}}

\usepackage[T1]{fontenc}

\usepackage[utf8]{inputenc}

\usepackage{microtype}

\newcommand{\overbar}[1]{\mkern 1.5mu\overline{\mkern-1.5mu#1\mkern-1.5mu}\mkern 1.5mu}

\title{Natural Language Satisfiability: Exploring the Problem Distribution and Evaluating Transformer-based Language Models}

\author{Tharindu Madusanka$^1$ \and Ian Pratt-Hartmann$^{1,2}$ \and Riza Batista-Navarro$^1$ \\
1. Department of Computer Science, University of Manchester;\\
2. Instytut Informatyki, Uniwersytet Opolski\\
}

\begin{document}
\maketitle
\begin{abstract}
Efforts to apply transformer-based language models (TLMs) to the problem of reasoning in natural language have enjoyed ever-increasing success in recent years. The most fundamental task in this area to which nearly all others can be reduced is that of determining satisfiability. However, from a logical point of view, satisfiability problems vary along various dimensions,  which may affect TLMs' ability to learn how to solve them. The problem instances of satisfiability in natural language can belong to different computational complexity classes depending on the language fragment in which they are expressed. Although prior research has explored the problem of natural language satisfiability, the above-mentioned point has not been discussed adequately. Hence, we investigate how problem instances from varying computational complexity classes and having different grammatical constructs impact TLMs' ability to learn rules of inference. Furthermore, to faithfully evaluate TLMs, we conduct an empirical study to explore the distribution of satisfiability problems. 
\end{abstract}

\section{Introduction}

The impressive performance of transformer-based language models (TLMs)  in natural language inference tasks \cite{devlin2018bert,yang2019xlnet,raffel2019exploring,liu2019roberta} has created a surge of interest in the development of linguistic and deductive reasoning benchmarks to evaluate these models \cite{richardson2020probing,geiger2018stress,Tafjord2021ProofWriterGI,yanaka-etal-2020-neural}. One such area of interest is the ability to recognise valid entailments, understood in a strictly logical sense \cite{clark2020transformers,richardson2021pushing}, where the inference does not depend on background knowledge and commonsense. 
The ability of TLMs to understand this type of entailment is indicative of their ability to learn rules of inference, understand the logical semantics of natural language and emulate complex algorithms.

\begin{figure*}
    \centering
    \includegraphics[width = 16cm]{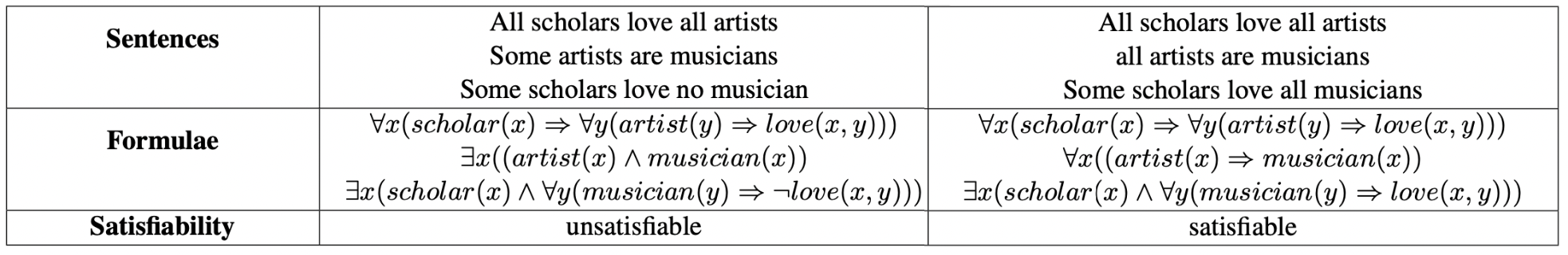}
    \caption{The table depicts two instances of a satisfiability problem; one is unsatisfiable while the other is satisfiable. In the first example, the first two formulae imply \(\forall x(scholar(x) \rightarrow \exists y(musicians(y) \land love(x,y)))\), while the last formula is \(\exists x (scholar(x) \land \forall y(musician(y) \rightarrow \neg love(x,y)))\)---a direct contradiction, hence unsatisfiable. In the 
    the second example (satisfiable), a structure \(\mathfrak{A}\) can be easily found that makes all formulae $True$: imagine, for instance, a world where scholars exist, all musicians are artists, and everyone loves everyone.}
    \label{fig:problem_sat}
\end{figure*}

Since the inferences we shall be concerned with do not depend on background knowledge or intuitive plausibility, they can be captured using the apparatus of formal logic.
Logicians usually find it convenient to reconstruct entailment in terms of satisfiability: a set of formulae $\Phi$ is \textit{satisfiable} if there is a structure (in 
the
model-theoretic sense) $\mathfrak{A}$ such that every 
formula of $\Phi$
is true in $\mathfrak{A}$. 
A set of formulae $\Phi$ \textit{entails} a formula $\psi$ just in case $\Phi \cup \{\neg \psi\}$ is not satisfiable. The same duality holds for natural language (which incorporates a concept of negation) just as it does for formal logic. Thus, we consider the following problem: given a collection of sentences 
expressed in a natural language such as English, determine whether it is---in a strictly logical sense---satisfiable. Figure \ref{fig:problem_sat} illustrates two instances of the satisfiability problem in English, one positive and one negative. 
Note that any method for determining satisfiability yields a method for recognising entailments and \textit{vice versa}.

We approach this problem in a controlled way. The sentences in Figure \ref{fig:problem_sat} feature only rudimentary grammatical resources: the determiners {\em some}, {\em no} and {\em all}, transitive verbs with unqualified noun-phrases as subjects and objects, and the copula {\em is}. It is well-known from formal logic that the computational complexity of the satisfiability problem for a logic depends on its expressive power.
For example, the satisfiability problem for propositional logic (known as SAT) is 
\NPTime-complete, but the corresponding problem for the two-variable fragment of first-order logic is 
\NExpTime-complete, while for the whole of first-order logic is r.e.-complete (i.e.~undecidable); for a survey, see Pratt-Hartmann ~\shortcite{pratt2023FOL}.
The same is true when it comes to fragments of natural languages \cite{pratt2003two,pratt2004fragments,pratt2006more}, as we explain presently. 
The question therefore arises as to how TLMs' ability to learn rules of inference is affected by the fragment of language within which
they operate.

In recent years, researchers have analysed the limitations of various neural approaches when solving satisfiability problems \cite{selsam2018learning,evans2018can,Cameron2020PredictingPS}, including natural language satisfiability problems \cite{richardson2021pushing}. However, in all cases, the problems are in propositional logic or its close relatives, and as such, the expressive power of the expressions or sentences is limited. Moreover, it restricts the problem space to be in a single computational complexity class, NP-complete. 
We overcome these limitations by generating sentences from various \textit{fragments} of English that are related to more varied fragments of first-order logic. By utilising language fragments of varying expressive power, we also provide an analysis of how computational complexity affects TLMs' ability to grasp the rules governing logical entailment. 

When investigating TLMs' ability to solve instances of natural language satisfiability problems, it is important to ensure that the training and test data sets include a sufficient number of challenging problem instances. The difficulty is that randomly constructed sets of formulae (or sentences) are, depending on the sampling parameters, in many cases easily seen either to be satisfiable or to be unsatisfiable, a situation dramatically illustrated in the case of propositional logic satisfiability \cite{selman1996generating,cook1997finding,mitchell1996some}, where challenging problems occur only when the ratio of clauses to propositional variables is close to a critical threshold, i.e., the so-called \emph{phase-change region}. In this region, where the probability of a randomly generated problem instance is close to 0.5, algorithms for determining satisfiability typically exhibit long running times. Therefore, to demonstrate that TLMs can learn rules of inference, we must ensure that they work on the hardest region of the target problem space. Indeed, recent work has shown some pitfalls associated with synthetic data due to under-sampling challenging problems \cite{shin2018synthetic,wu2021reascan}. Consequently, we conduct an empirical study to determine, for each of the language fragments investigated, the relevant 
``phase-change'' region where the challenging problem instances are to be found. 

The contributions of this paper are as follows. (1) We identify a number of fragments of English featuring a range of logico-syntactic constructions involving determiners, the copula, transitive verbs, relative clauses and bound-variable anaphora. (2) We empirically determine the phase-change region for these fragments and construct data sets containing instances of the satisfiability problems
sampled from these regions. (3) We investigate the ability of transformer-based language models to solve the satisfiability problem for the investigated fragments and carry out a systematic analysis of how the underlying computational complexity of the satisfiability problem correlates with TLMs' ability to grasp the relevant inferential principles. (4) Furthermore, we explore the proficiency of TLMs in solving instances of the satisfiability problem within a zero-shot context. 
To the best of our knowledge, this investigation represents the first attempt to probe the behaviour of TLMs in zero-shot scenarios with respect to their efficacy in solving satisfiability problems.

\section{Related Work}
Our work closely follows the literature on identifying strengths and weaknesses of neural approaches, including transformer-based language models on deductive reasoning tasks \cite{mccoy-etal-2019-right,glockner-etal-2018-breaking,lin-etal-2019-reasoning}. Neural networks, particularly graph neural networks (GNN), have been utilised to solve satisfiability problem instances \cite{Xu2020What,cameron2020predicting}. As mentioned in the introduction, various studies have extended that work to natural language satisfiability, utilizing TLMs instead of GNNs \cite{richardson2021pushing}. However, the sentences considered fail to exercise the full range of inference patterns licenced by commonly encountered grammatical constructions. Moreover, in each case, the underlying satisfiability problems were derived from that for propositional logic (i.e.~SAT), thus, confining attention to a single problem type. Indeed in the case of natural language satisfiability, the sentences are sometimes hardly natural-sounding. For example, the Grounded Rule Language (GRL) fragment introduced in Richardson and Sabharwal \shortcite{richardson2021pushing} includes sentences such as \textit{``If carrot and not steak then apples''}.

Our work can also be viewed as an attempt to identify various limitations that affect TLMs' ability to emulate algorithms. There have been numerous studies of TLMs' ability to solve algorithmic tasks, including SAT-solving \cite{selsam2018learning,Narodytska2020In}, semantic parsing \cite{he-choi-flairs-2020,kamath2019a}, model-checking \cite{madusanka-etal-2023-identifying,madusanka-etal-2023-quantifiers}, theorem proving \cite{weber-etal-2019-nlprolog,DBLP:conf/aaai/MinerviniBR0G20,saha-etal-2020-prover,welleck2021naturalproofs} etc. We extend this work to a family of related problems spanning multiple computational complexity classes, thus, providing a breakdown of the impact of computational complexity on TLMs' ability to solve algorithmically challenging problems. 

\section{Methodology}

\subsection{Language Fragments}
\label{sec:languageFragments}
By a \textit{natural language fragment}, we understand a set of sentences in some natural language equipped with truth-conditional
semantics in terms of which the dual notions of validity and satisfiability can be defined. Consider, for example, the set of sentences of the forms
\begin{center}
\begin{minipage}{4cm}
\begin{tabbing}
{Every $p$ is a $q$} \qquad  \=  {Some $p$ is a $q$}\\
{No $p$ is a $q$}     \>         {Some $p$ is not a $q$},
\end{tabbing}
\end{minipage}
\end{center}
where $p$ and $q$ range over common (count) nouns such as {\em artist}, {\em beekeeper}, {\em carpenter} \ldots. Their semantics can be given by the familiar 
translation into first-order logic over a non-logical vocabulary (of unary predicates) corresponding to the schematic variables in question:
\begin{center}
\begin{minipage}{4cm}
\begin{tabbing}
$\forall x (p(x) \rightarrow q(x))$ \qquad  \=  $\exists x (p(x) \wedge q(x))$\\
$\forall x (p(x) \rightarrow \neg q(x))$    \>  $\exists x (p(x) \wedge \neg q(x))$.
\end{tabbing}
\end{minipage}
\end{center}
These semantics assume that universal sentences have no existential import (i.e.~\textit{Every $p$ is a $q$} does not entail that there \textit{are} any $p$'s); otherwise, however, they are 
uncontentious, yielding a faithful reconstruction of the notions of validity (of an argument) and satisfiability (of a set of sentences) via the usual model-theoretic definitions. In this way, we have defined a fragment of English---one corresponding, modulo translation, to the classical syllogistic of Aristotle's {\em Prior Analytics}, Book A~\cite{aristotle}. 

\begin{figure*}
    \centering
    \includegraphics[width = 16cm]{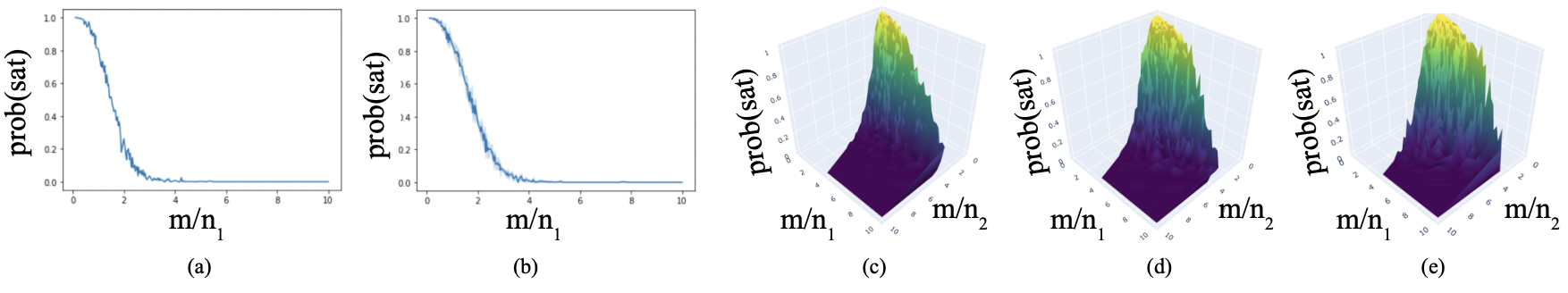}
    \caption{The probability of satisfiability for sets of sentences in the language fragments: (a) $\mathcal{S}$, (b) $\mathcal{W}$, (c) $\mathcal{V}$, (d) $\mathcal{Z}$ and (e) $\mathcal{A}$. Here, $m$ denotes the number of clauses, and $n_1$, and $n_2$ the number of common nouns and transitive verbs, respectively, in the sampled vocabulary. }
    \label{fig:prob_sat}
\end{figure*}

Generalising, say that a \textit{sentence template} is a sentence of some natural language in which certain open-class words have been replaced by schematic variables; and say that a \textit{formula template} of some logic is a formula (with no free variables) in which the same schematic variables are treated as elements of the non-logical signature, appropriately typed. Then a simple way to define a natural language fragment is by means of a finite list of sentence templates paired with corresponding formula templates giving their semantics in a way judged appropriate by competent speakers. We consider five English fragments in the sequel, all defined in this way: 
(i) the fragment $\mathcal{S}$, a version of the syllogistic in which---for the sake of logical uniformity---we additionally allow negation at the subject (e.g.~{\em Some non-artist is not a beekeeper});
(ii) the fragment $\mathcal{W}$, which adds relative clauses to the subjects of sentences in $\mathcal{S}$ (e.g.~{\em Every artist \textit{who is not a musician} is a writer}); 
(iii) the fragment $\mathcal{V}$, which extends $\mathcal{S}$ with main clauses featuring transitive verbs and a quantifying determiner (e.g.~{\em Every carpenter \textit{admires some} writer});
(iv) the fragment $\mathcal{Z}$, which extends $\mathcal{W}$ with relative clauses featuring transitive transitive verbs
(e.g.~{\em Every carpenter who admires some writer is an electrician}); and
(v) the fragment $\mathcal{A}$, which extends $\mathcal{Z}$ with certain sentence templates involving bound-variable anaphora (e.g.~{\em Some artist hates no beekeeper who admires him}). The precise sets of sentence templates defining these fragments are presented at the end of this paper. Since the relevant sentence templates 
correspond to sets of formula templates of first-order logic, 
our five language fragments may, to all intents and purposes, be identified with the corresponding fragments of first-order logic; hence we alternate freely between sentences and their logical translations, as the context requires.

If $\mathcal{L}$ is a natural language fragment,  we denote by $Sat(\mathcal{L})$ the problem:
\begin{center}
\begin{minipage}{10cm}
\begin{tabbing}
Given: a finite set $\Phi$ of sentences in $\mathcal{L}$,\\
Return: Y if $\Phi$ is satisfiable; N otherwise.
\end{tabbing}
\end{minipage}
\end{center}

It can be shown that $\Sat(\mathcal{S})$ is \NLogSpace-complete
and $\Sat(\mathcal{V})$  is \ExpTime-complete~\cite[Theorems~4.11 and~6.3]{pratt2009logics}.
It is completely routine to show that $\Sat(\mathcal{W})$ is \NPTime-complete. With a little effort, one can show that~$\Sat(\mathcal{Z})$ is 
also \ExpTime-complete, using essentially the same techniques as for $\Sat(\mathcal{V})$. Finally, $\Sat(\mathcal{A})$
is \NExpTime-complete: membership in \mbox{\NExpTime{}} follows from the fact that all the corresponding formula 
templates are in the two-variable fragment of first-order logic~\cite{gkv97}; \NExpTime-hardness follows
by an almost identical argument to that given for the rather larger English 
fragment E2V in~\cite{pratt2003two}. The proof, along with a detailed complexity-theoretic analysis of the language fragments discussed, are depicted in Appendix \ref{sec:appendix_complexity}.
Given these results, it is natural to ask whether language models have greater success in learning to recognise valid entailments (or, dually: satisfiability of sets of sentences) 
for the computationally easier fragments, or whether training on easier fragments aids in recognising entailments in harder ones.



\subsection{Identifying the phase change region}

It is important to realise that the worst-case complexity bounds for satisfiability problems are not necessarily representative of randomly generated problem instances, a phenomenon that is well understood in the case of propositional logic. Suppose that $k$ is a positive integer and that a $k$-{\em clause} is a disjunction of $k$ or fewer proposition letters or negated proposition letters. The $k$-SAT problem is the following: given a finite set of clauses $\Gamma$, return 1 if $\Gamma$ is satisfiable (i.e.~if there is a truth-value assignment making all the clauses in $\Gamma$ \textit{true}), and 0 otherwise. This problem is \NPTime-complete for all $k \geq 3$. However, under certain conditions, it is trivial to decide random instances of these problems with high probability. Fixing $k=3$, consider a randomly generated instance $\Gamma$ of 3-SAT consisting of
 $m$ clauses featuring $n$ proposition letters $p_1, \dots, p_n$ (and their negations). If the ratio $\frac{m}{n}$ is large, we have a highly constrained problem with few degrees of freedom, so the probability of satisfiability is close to zero; if, on the other hand, the ratio $\frac{m}{n}$ is small,
 the probability of satisfiability is close to unity. In either case, it is trivial for any algorithm to achieve high performance 
 when predicting the correct answer. Only for values in a relatively narrow range of $\frac{m}{n}$, commonly
 referred to as the {\em phase-change region}, is the problem challenging, with this range
 (centred on a value close to 4.17) narrowing as $n$ increases~\cite{selman1996generating,mitchell1996some}.

A similar phenomenon is observed for the satisfiability problems for natural language fragments studied here: the satisfiability of a given set of sentences may in many cases be easily determined with high probability by measuring the number of degrees of freedom in the given instance. Consider, for example, the problems Sat($\mathcal{S}$) and Sat($\mathcal{W}$). Any instance of these problems is characterised by the number $n_1$ of common nouns featured, and the number $m$ of sentences involved. We find that the probability of satisfiability for randomly generated 
instances is close to 0.5 only when the ratio $\alpha = \frac{m}{n_1}$ is in a relatively narrow band. For the other
fragments considered in this study, which feature transitive verbs as well as common nouns, we must consider not only the ratio $\alpha$ but also the ratio $\beta = \frac{m}{n_2}$, where $n_2$ is the number of transitive verbs in the given problem instance. Again, we find that 
the probability of satisfiability for a randomly generated problem is close to 0.5 only for pairs of values $(\alpha, \beta)$ lying in a relatively constrained region. We continue to refer to this region, informally, as the \textit{phase-change region}: notice that this is a region in 2-dimensional space. 
It is important to remember that the shape and location of the phase-change region depend on the fragment considered; however, in all cases, the gradients involved are observed to become sharper with increasing $n$. In assessing the ability of TLMs to learn to solve the satisfiability problem for the fragments considered in this paper, we must make sure to construct data sets involving only challenging instances, namely, those selected from the phase-change region: prowess at solving cases chosen uniformly over parameter space is a poor test of a system's grasp of logical principles. For definiteness, we take the phase-change region
for a fragment $\mathcal{L}$, denoted 
$\lambda_\mathcal{L}$,
to be the set of input parameters for which the probability of satisfiability is [0.35, 0.65]. This set can be determined empirically by random sampling, much as in the original studies of
3-SAT.
The variation of probability of satisfiability with other factors is outlined in Appendix \ref{sec:appendix_phase_change_region}.

\subsection{Data Construction}

For each of the language fragments $\mathcal{L}$ considered above, we constructed a \textit{data set} with which to fine-tune TLMs. Each data set is a set 
of \textit{data points}. Each data point is a pair consisting of a set of sentences $\Phi$ from $\mathcal{L}$, and a label (Y or N) indicating whether $\Phi$ is satisfiable. To generate a single sentence of $\Phi$, we select at random one of the 
sentence templates defining $\mathcal{L}$ and instantiate its schematic variables by uniform sampling from a collection of $n_1$ common nouns and (for the applicable fragments) $n_2$ transitive verbs; repeating this whole process $m$ times then yields a set $\Phi$ of cardinality $m$. The label determining the satisfiability of $\Phi$ is determined by applying a theorem prover (in our case, \texttt{Z3}) to the corresponding set of formulae of first-order logic as given by the formula templates. We construct data points for various values of $m$, $n_1$ and (where appropriate)
$n_2$, taking care to employ only those combinations within the critical region, $\lambda_\mathcal{L}$. As explained above, $\lambda_\mathcal{L}$ has been determined empirically. 

For the fragments $\mathcal{S}$ and $\mathcal{W}$, we sample values of $n_1$ from the range $[n_1^{min}, n_1^{max}]$, where
$n_1^{min}$ = 6, and $n_1^{max}$ = 16.
For the fragments $\mathcal{V}$, $\mathcal{Z}$, and $\mathcal{A}$, we sample values of $n_1$ from the range $[n_1^{min}, n_1^{max}]$, where
$n_1^{min}$ = 3, $n_1^{max}$ = 8, and we sample values of $n_2$ from the range $[n_2^{min}, n_2^{max}]$, where $n_2^{min}$ = 3 and $n_2^{max}$ = 8.
In each case, we generate a set of values of values $m$ for which $\alpha = m/n_1$ and (for the appropriate fragments) $\beta = m/n_2$ lie in $\lambda_{\mathcal{L}}$.
The entire protocol for generating data sets is given in Algorithm~\ref{alg:one}. We use the theorem prover \texttt{Z3} to determine the satisfiability of the generated set of 
formulae, $\Phi = \{\phi_1, \dots, \phi_m\}$.
For each language fragment, we construct a training set with 120K data points, an evaluation set with 10K data points and a test set with 10K data points to fine-tune and evaluate TLMs. 
We note that the above experimental setup can be directly adapted to any language fragment.
\begin{algorithm}
\SetAlgoLined
 \textbf{Input :} Language Fragment \(\mathcal{L}\); phase-change region $\lambda_\mathcal{L}$ (for $\mathcal{L}$); vocabulary of unary predicates \(\mathcal{U}\); vocabulary of binary predicates $\mathcal{V}$; range for number of unary predicates $[n_1^{min}, n_1^{max}]$; range 
 for number of binary predicates $[n_2^{min}, n_2^{max}]$.\\
\textbf{Output :} natural language satisfiability dataset \(\mathcal{D}\).\\
\begin{algorithmic}[1]
    \STATE \(D \gets \{\}\)
    \REPEAT 
        \STATE \(m, n_1, n_2 \gets \) sample \(m, n_1, n_2\) such that $n_1^{min} \leq n_1 \leq
        n_1^{max}$, $
        n_2^{min} \leq n_2 \leq n_2^{max}$ and $(\frac{m}{n_1},\frac{m}{n_2}) \in \lambda_\mathcal{L}$
        \STATE \(\mathcal{U}^*,\mathcal{V}^*  \gets\) sample \(n_1\) unary predicates and \(n_2\) binary predicates from \(\mathcal{U}\) and $\mathcal{V}$ respectively, \(|\mathcal{U}^*| = n_1\) and \(|\mathcal{V}^*| = n_2\)
        \FOR{\(i = 1\) to \(m\)}
            \STATE \(t_i \gets\) randomly sample template from language fragment $\mathcal{L}$ 
            \STATE $s_i \gets$ substitute predicates from $\mathcal{U}^*$ and 
            $\mathcal{V}^*$
            for schematic variables in $t_i$
            \STATE $\phi_i \gets$ translate \(s_i\) to a first-order logic formula
        \ENDFOR
        \STATE \(\ell \gets\) \texttt{SAT-solver}\((\{\phi_1, ..., \phi_m\})\)
        \STATE \(\mathcal{D} \gets \mathcal{D} \cup \{\langle \ell, \{s_{1}, ...,  s_{m}\}\rangle\}\)

    \UNTIL{\textit{stop condition is met}}
\caption{Data Construction - Natural language satisfiability}\label{alg:one}    
\end{algorithmic}   
\end{algorithm}

For evaluating TLMs in zero-shot settings we set  $n_1^{min}$ = 5, and $n_1^{max}$ = 10 for fragments $\mathcal{S}$ and $\mathcal{W}$ and $n_1^{min}$ = 3, $n_1^{max}$ = 5, $n_2^{min}$ = 2 and $n_2^{max}$ = 5 for fragments $\mathcal{V}$, $\mathcal{Z}$, and $\mathcal{A}$. 
We then construct an additional 1200 problem instances for each language fragment and formulate the prompt using the constructed problem instances employing a template-based approach. The exact template is depicted in Appendix \ref{sec:zeroshot_templates}. Further details regarding the data sets are given in Appendix \ref{sec:appendix_training}.



\section{Experimental Setup}

The objective of the TLM is to approximate $\Omega$ given training instances $\mathcal{D}_{tr} = \{(P^{(d)}, \Omega(P))^{(d)}\}^{|D_{tr}|}_{(d)}$. Since the $\Omega(P)$ is a binary label, the problem can be modelled as a binary classification problem, where the label is set to be $1$ or $0$ based on satisfiability. 

\subsection{Transformer-based language models}
\label{sec:transformerModels}

To examine TLMs' ability to solve hard instances of natural language satisfiability problems, and investigate their capability to learn rules of inference, we fine-tuned two well-known TLMs which have a proven track record of solving textual entailment problems\footnote{Link to the Dataset and Code: \url{https://github.com/iTharindu/nl-sat}}.

\paragraph{T5.}
Following the work done by Richardson and Sabharwal \shortcite{richardson2021pushing} and Tafjord et al. \shortcite{Tafjord2021ProofWriterGI} on logical reasoning in natural language, we primarily centred our experiments around the text-to-text transformer or  T5 \cite{raffel2019exploring}. The T5 model employs a unified text-to-text format where all inputs and outputs are textual strings, in contrast to BERT-styled models. We utilised the \texttt{T5-large} model, which has around 700M parameters. 

\paragraph{DeBERTa-v3.}
Due to the recent success of the DeBERTa-v3 model \cite{he2021debertav3} in solving natural language inference tasks, we decided to use it as a baseline model. The DeBERTa architecture improves upon the BERT and RoBERTa models using a disentangled attention mechanism and enhanced mask 
decoder,
and version 3 further improves the architecture by utilising an ELECTRA-style pre-training with Gradient Disentangled Embedding Sharing. We employ the \texttt{DeBERTa-v3-large} model with around 304M parameters. 

Each of the TLMs is fined-tuned by reducing the binary cross entropy loss over the target using Adam optimiser \cite{DBLP:journals/corr/KingmaB14} and we used the HuggingFace \cite{wolf2019huggingface} implementation when experimenting with the above-mentioned TLMs. 
A detailed description of the fine-tuning process is described in Appendix \ref{sec:appendix_training}.

To investigate TLMs' ability to solve satisfiability problems in \textit{zero-shot} settings, we employed three well-known models.

\paragraph{GPT.} 
Due to the recent success of ChatGPT and GPT-4 solving many natural language processing tasks in zero-shot settings, we employed them in a similar context \cite{bang2023multitask,openai2023gpt4}. Both ChatGPT and GPT-4 models are trained to follow human instructions utilising Reinforcement Learning from Human Feedback (RLHF). 

\paragraph{LLaMa.}
To better compare the effect of model size and architecture, we also utilise the \texttt{LLaMA-2-chat-70B} model in zero-shot settings. LLaMa-2 achieves comparable performance with state-of-the-art language models such as ChatGPT and PALM \cite{touvron2023llama}. Similar to ChatGPT and GPT-4, the LLaMa-2-chat model also employs RLHF. 

\subsection{Proposed Dataset and Evaluation}

To evaluate the ability of TLMs to solve natural language satisfiability problem instances, we designed several experiments. Firstly, to answer the questions, ``Can TLMs solve hard natural language satisfiability problems?'' and ``How does computational complexity affect TLMs' ability to perform a logical reasoning task?'', we trained and evaluated the TLMs mentioned in Sec.~\ref{sec:transformerModels} against each of the language fragments introduced in Sec.~\ref{sec:languageFragments} (see Table \ref{table:experiment 01}). Secondly, to answer the question ``Do the computationally simpler language fragments help TLMs learn rules inference in complex ones?'', we trained the same TLMs using a dataset that comprises problem instances from all language fragments (see Table \ref{table:experiment 02}). We employed two variants of this joint dataset, one with 600K data points with 120K data points from each fragment and the other with 120K data points with 24K data points from each fragment. Thirdly, to answer the question ``Do 
TLMs show the ability to generalise and show scale-invariance properties?'', we evaluated the same TLMs against a dataset containing more predicates (variables) than that of the training set (see Table \ref{table:experiment 03}).
This provided clarity into TLMs' ability to learn rules of inference from natural language satisfiability problem instances. Finally, to answer the question, "What factors affect TLMs' ability to solve Satisfiability problem instances?", we evaluated large-scale transformer-based language models in zero-shot settings. We employ a much smaller dataset in this evaluation and the dataset description is detailed in Appendix \ref{sec:appendix_training}. The datasets we constructed were balanced with an equivalent number of satisfiable and unsatisfiable instances, and, consequently, we used accuracy as the evaluation metric. 

\section{Results and Discussion}

\textbf{The ability of TLMs to solve natural language satisfiability problems is affected by the underlying computational complexity of the problem, as
determined by the language fragment in question.} As shown in Table \ref{table:experiment 01}, the performance of TLMs considered declines as 
the computational complexity class increases. It also can be observed that descent is steeper from $\mathcal{S}$ to $\mathcal{W}$ and more gradual henceforward. We posit two reasons for this phenomenon. Firstly, the problems that are in \textsc{NLogSpace} can be solved by the detection of relatively simple configurations (forbidden configuration of relatively low length in case the sentences are unsatisfiable) that TLMs can easily learn to recognise, while those that are \textsc{NPTime}-hard (or harder) are characterised by more intricate structures. Secondly, the problems in the simpler fragments have fewer tokens in their input sentences compared to those in the more complex language fragments; and TLMs (like other neural approaches) find it more challenging to process long dependencies than short ones \cite{chen2020differentiable,Beltagy2020Longformer}. However, computational complexity is not the only factor that influences the behaviour of the TLMs when performing logical inference tasks, as evidenced by the difference in performance when predicting the satisfiability of problem instances in $\mathcal{V}$ and $\mathcal{Z}$, both of which are 
\ExpTime-complete. The two fragments contain the same language properties, such as relative clauses and transitive verbs, but $\mathcal{Z}$ has sentences that retain both of those properties together while $\mathcal{V}$ does not. Thus, some sentences in $\mathcal{Z}$ are intrinsically more linguistically complicated than those in $\mathcal{V}$. As the neural approaches we employed are pre-trained language models that were trained on large language corpora, the linguistic complexity of the input has a noticeable effect on their performance.

\begin{table}
    \centering
    \begin{tabular}{|c|c|c|} \hline 
         \textbf{Fragment}&  \makecell{T5-large} & \makecell{DeBERTa-v3-large} \\ \hline 
         \makecell{\textbf{$\mathcal{S}$}} 
&  88.3& 89.0\\ \hline 
         \makecell{\textbf{$\mathcal{W}$}} 
&  80.7& 78.6\\ \hline 
         \makecell{\textbf{$\mathcal{V}$}} 
&  79.5& 77.2\\ \hline 
         \makecell{\textbf{$\mathcal{Z}$}} 
&  75.1& 75.2\\ \hline 
         \makecell{\textbf{$\mathcal{A}$}}   &  70.1& 70.9\\ \hline
    \end{tabular}
    \caption{Accuracy of TLMs (T5-large and DeBERTa-V3) when fine-tuned and evaluated across the fragments $\mathcal{S}$, $\mathcal{W}$, $\mathcal{V}$,$\mathcal{Z}$, and $\mathcal{A}$.}
    \label{table:experiment 01}
\end{table}
\begin{table}
    \centering
    \begin{tabular}{|c|c|c|} \hline 
         \textbf{Fragment}&  \makecell{T5-large\textsubscript{600k}} & \makecell{T5-large\textsubscript{120k}} \\ \hline 
         \makecell{\textbf{$\mathcal{S}$}} &  86.7& 74.1\\ \hline 
         \makecell{\textbf{$\mathcal{W}$}} &  85.0& 73.7\\ \hline 
         \makecell{\textbf{$\mathcal{V}$}} &  83.7& 69.5\\ \hline 
         \makecell{\textbf{$\mathcal{Z}$}} &  83.3& 68.6\\ \hline 
         \makecell{\textbf{$\mathcal{A}$}}   &  82.2& 68.0\\ \hline
    \end{tabular}
    \caption{Accuracy of TLMs (T5-large) when trained on a dataset containing problem instances of all fragments; the test accuracies are broken down based on the language fragment, ($\mathcal{S}$, $\mathcal{W}$, $\mathcal{V}$,$\mathcal{Z}$, and $\mathcal{A}$). T5-large\textsubscript{600k}, and T5-large\textsubscript{120k} indicate that the training set contains 600K and 120K data points respectively.}
    \label{table:experiment 02}
\end{table}

\textbf{Provided adequate data, learning to solve satisfiability problem instances of simpler fragments can help TLMs learn to solve that of complex ones.} When the dataset contains problem instances from all fragments, as shown
in Table \ref{table:experiment 02}, the accuracy value for complex fragments increases when trained with an adequate number of data points (accuracy values for 600K). We hypothesise two reasons for this; first, fragments whose computational complexity for solving satisfiability problems is high have sentences from the fragments 
such as $\mathcal{S}$, which are much simpler; 
second, even in complex language fragments such as $\mathcal{A}$, the inference is heavily governed by the sentences belonging to simpler fragments (such as involved in proofs when the set of sentences is unsatisfiable). Thus, having problem instances belonging only to simpler fragments helps TLMs decode reasoning patterns caused by sentences belonging to them. However, this raised a concern as to the true difficulty of the underlying dataset, even when sampled from the phase-change region. When the number of data points is low, say 120K data points, the performance of the TLMs decreases for all fragments. We consider 24K data points from a single language fragment inadequate to learn any meaningful representation required for determining satisfiability. This is somewhat dissimilar to the results yielded by the computationally simpler rule reasoning problem of model-checking with natural language \cite{madusanka-etal-2023-identifying}.

\begin{table}
\centering
  \begin{tabular}{|m{1.8cm}|m{2.2cm}|m{2.2cm}|} 
  \hline
  \makecell{\textbf{Fragment}} & \makecell{\textbf{$n_1 + n_2 \leq 20$}} & \makecell{\makecell{\textbf{$n_1 + n_2 \ge 20$}} } \\ \hline

\makecell{$\mathcal{S}$} & \makecell{75.0} & \makecell{71.3}\\ \hline
\makecell{$\mathcal{W}$} & \makecell{75.3} & \makecell{71.2}\\ \hline
\makecell{$\mathcal{V}$} & \makecell{69.0} & \makecell{68.9}\\ \hline
\makecell{$\mathcal{Z}$} & \makecell{68.9}& \makecell{68.5}\\ \hline
\makecell{$\mathcal{A}$} & \makecell{64.3}& \makecell{58.0}\\ \hline

  \end{tabular}
  \newline
  \caption{TLM's (T5-large) ability to generalise to harder problems. The models were trained on problems with $ 6 \leq n_1 + n_2 \leq 16$ and evaluated against problems with $n_1 + n_2 \geq 16$.}
  \label{table:experiment 03}
\end{table}

\textbf{TLMs failed to generalise and learn rules of inference required to solve satisfiability problem instances.} 
We investigated whether TLMs learn to apply rules of inference and comprehend the underlying algorithm associated with the satisfiability-solving task by evaluating them against a dataset containing more predicates (thus more clauses) than that of the training test. As depicted in Table \ref{table:experiment 03}, TLMs failed to generalise, and accuracy across all fragments decreased regardless of the computational complexity,
suggesting TLMs do not systematically generalise. This hypothesis is 
reinforced by the evident failure of TLMs to generalise even for fragment $\mathcal{S}$, for which they exhibited a good accuracy level. Prior literature on the systematic generalisation of neural approaches \cite{Lake2017GeneralizationWS,goodwin-etal-2020-probing} also corroborates our finding that neural networks generally do not generalise well outside the training distribution. Moreover, experiments conducted by Richardson and Sabharwal \shortcite{richardson2021pushing} on scale invariance yielded equivalent results.

\begin{figure}
    \centering
    \includegraphics[width=0.6\linewidth]{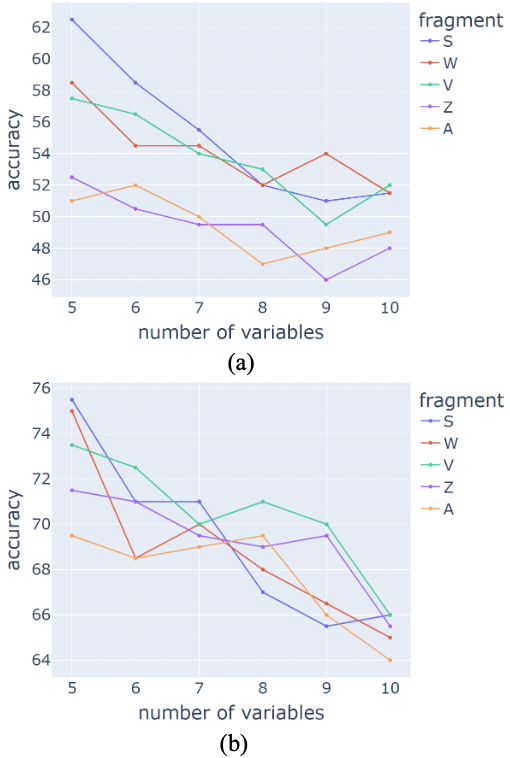}
    \caption{Variation of accuracy for (a) ChatGPT and (b) GPT-4 for language fragments, $\mathcal{S}$, $\mathcal{W}$, $\mathcal{V}$,$\mathcal{Z}$, and $\mathcal{A}$ for different number of variables}
    \label{fig:zero-shot}
\end{figure}

\textbf{Pre-trained transformer-based language models do not achieve adequate performance when solving even the simplest satisfiability problems in zero-shot settings, and factors such as model architecture, number of variables and exact prompt can affect models' ability to follow instructions.} The GPT-4 model achieved the best performance among the models we consider and is the only model whose results are reasonably higher than random guessing. As depicted in Figure \ref{fig:zero-shot}, notably, the GPT-4 model outperformed ChatGPT by a considerable margin for all language fragments, which is in line with prior evaluations on algorithmic tasks such as coding and answering Mathematics questions \cite{openai2023gpt4}. Another notable observation is that the effect of the computational complexity of the language fragments when solving natural language satisfiability problem instances is considerably 
lower
compared to fine-tuned models. 

A deeper analysis of the answers generated also shows that GPT-4 is better at explicitly following the instructions. For example, if the prompt indicates only to generate ``satisfiable'' or ``unsatisfiable'', and asks to generate nothing else, the GPT-4 model follows that instruction explicitly more often than the other models.  Conversely, our experimentation with LLaMa-2 reveals a contrary trend; for most problem instances, the models do not follow the instructions and often generate something other than ``satisfiable'' or ``unsatisfiable''.

When we introduced a modification to the prompt by incorporating an illustrative example and requesting the generation of ``True" or ``False" instead of ``satisfiable'' or ``unsatisfiable'', the model consistently produced responses of True or False. We hypothesise the model's failure to adhere to the instructions of the original prompt can be attributed to two factors. Firstly, the terms ``satisfiable'' and ``unsatisfiable'' are relatively infrequent in the text data the language model has been trained on. Consequently, from a probabilistic standpoint, the likelihood of these specific terms being the most probable next tokens is low. Therefore, the model often generates something other than  ``satisfiable'' or ``unsatisfiable''. Secondly, the inclusion of an illustrative example within the modified prompt provides the model with an unambiguous context regarding the desired output. However, even in this scenario, the model's performance is equivalent to random guessing. 

Unsurprisingly, when the number of variables increases the accuracy of GPT models decreases. We posit two reasons for this observed phenomenon, 
Firstly, from a logical perspective, the increment in the number of variables increases the difficulty of solving the satisfiability problem. As the number of variables grows, the search space expands, rendering the task of finding a valid assignment more challenging.
Secondly, the increment of the number of variables increases the total number of tokens within a given problem instance. This increase in token count compels the model to grapple with longer-term dependencies in the input. This gradual decrement along with the recurring failure of fine-tuned TLMs to generalise, leads to the hypothesis that, notwithstanding their impressive performance, state-of-the-art transformer models are still far from learning the rules of inference underlying logical reasoning tasks, and the algorithms required to apply them.

\section{Conclusion}

We have investigated transformer-based language models' ability to solve instances of natural language problems belonging to different language fragments with varying computational complexity. Our investigation demonstrated that the computational complexity of the fragment has a noticeable effect on TLMs' ability to perform a logical inferential task. Even in a simpler language fragment for which the transformer models achieved a reasonable performance, the model failed to adjust to distribution shift and generalise beyond its training distribution. Thus, we posit that TLMs do not reliably disentangle the patterns that exist within the dataset and the rules of inference needed when determining satisfiability. Therefore, it is imperative to acknowledge that a considerable body of research remains to be conducted in order to enhance the capacity of these models to comprehend rules of inference. We also acknowledge that the generalisation aspect of TLMs may need a deeper analysis, which we leave for future work.

\section*{Limitations}
The empirical study presented in this paper exhibits two principal limitations. We explored several TLMs with different architectures that are in line with prior work and studied how computational complexity and other linguistic factors affect those models when learning an inferential task. However, due to the empirical nature of the research, there could exist a language model architecture whose behaviour deviates from the claims made in this paper. Similarly, the language fragments employed in this literature are not the only fragments in existence and, as such, would affect the comprehensiveness of the discussion. 

Moreover, when constructing the dataset, we sampled problem instances from the phase-change region, and it was observed that the theorem prover took significantly more time to determine the satisfiability when data points were sampled from the phase-change region as opposed to random sampling. However, it does not provide a guarantee as to whether all instances are non-trivial, and it is probable that some problem instances are still trivial to solve. 

\section*{Acknowledgements}
Ian Pratt-Hartmann was supported by the Polish NCN, grant 2018/31/B/ST6/03662.
The authors would like to acknowledge the use of the Computational Shared Facility at the University of Manchester and the Department of Computer Science at the University for funding this research. We also thank the anonymous reviewers from the ARR December 2023 cycle for
their valuable feedback.


\appendix

\section{Appendix: Definition of language fragments}
\label{sec:templates}

The sentence templates defining $\mathcal{S}$, together with the corresponding formula templates, are as follows:
\begin{center}
\begin{minipage}{7cm}
\begin{tabbing}
Every (non-) $p$ is a $q$ \qquad \qquad No (non-) $p$ is a $q$\\
Some (non-) $p$ is (not) a $q$.
\end{tabbing} 
\end{minipage}
\end{center}
\noindent 
We can define the first-order formulae for giving semantics for the above sentence templates as follows:
\begin{center}
\begin{minipage}{7cm}
\begin{tabbing}
$\forall x (\pm p(x) \rightarrow \pm q(x))$ \\
$\exists x (\pm p(x) \wedge \pm q(x))$,
\end{tabbing} 
\end{minipage}
\end{center}
where $\pm \psi$ is either $\psi$ or $\neg \psi$. 

The sentence templates for $\mathcal{W}$  are those of $\mathcal{S}$ together with
\begin{center}
\begin{minipage}{7cm}
\begin{tabbing}
Every (non-) $o$ who is (not) a $p$ is a $q$\\
No (non-) $o$ who is (not) a $p$ is a $q$\\
Some (non-) $o$ who is (not) a $p$ is (not) a $q$.
\end{tabbing} 
\end{minipage}
\end{center}
\noindent 
We can define the first-order formulae for giving semantics for the above sentence templates as follows:
\begin{center}
\begin{minipage}{7cm}
\begin{tabbing}
$\forall x ((\pm o(x) \wedge\pm p(x)) \rightarrow \pm q(x))$ \\
$\exists x (\pm o(x) \pm p(x) \wedge \pm q(x))$.
\end{tabbing} 
\end{minipage}
\end{center}

The sentence templates for $\mathcal{V}$  are those of $\mathcal{W}$ together with
\begin{center}
\begin{minipage}{7cm}
\begin{tabbing}
Some/every (non-) $p$ $r$'s some/every (non-) $q$\\
No (non-) $p$ $r$'s any/every (non-) $q$\\
Some (non-) $p$ does not $r$ any/every (non-) $q$.
\end{tabbing} 
\end{minipage}
\end{center}
We can define the first-order formulae for giving semantics for the above sentence templates as follows,
\begin{center}
    \begin{minipage}{7cm}
        \begin{tabbing}
            $\forall x(\pm p(x)\rightarrow \forall y(\pm q(y) \rightarrow \pm r(x, y)))$ \\
            $\forall x(\pm p(x)\rightarrow \exists y(\pm q(y) \wedge \pm r(x, y)))$ \\
            $\exists x(\pm p(x)\wedge \forall y(\pm q(y) \rightarrow \pm r(x, y)))$ \\
            $\exists x(\pm p(x)\wedge \exists y(\pm q(y) \wedge \pm r(x, y)))$. 
        \end{tabbing}
    \end{minipage}
\end{center}
\noindent 

The sentence templates for $\mathcal{Z}$ are those of $\mathcal{V}$ together with
\begin{center}
\begin{minipage}{7cm}
\begin{tabbing}
Some (non-) $o$ \=who $r$'s/does not $r$ \\ 
\> some/every (non-) $p$ is (not) a $q$\\
Every (non-) $o$ who $r$'s/does not $r$  \\ 
\> some/every (non-) $p$ is a $q$\\
No (non-) $o$ who $r$'s/does not $r$ \\ 
\> any/every (non-) $p$ is a $q$.
\end{tabbing} 
\end{minipage}
\end{center}
We can define the first-order formulae for giving semantics for the above sentence templates as follows:
\begin{center}
    \begin{minipage}{7cm}
        \begin{tabbing}
            $\forall x (\pm o(x) \wedge \forall y(\pm p(y) \rightarrow \pm r(x,y)) \rightarrow \pm q(x))$ \\
            $\forall x (\pm o(x) \wedge \forall y(\pm p(y) \wedge \pm r(x,y)) \rightarrow \pm q(x))$ \\
            $\exists x (\pm o(x) \wedge \forall y(\pm p(y) \rightarrow \pm r(x,y)) \wedge \pm q(x))$ \\
            $\exists x (\pm o(x) \wedge \forall y(\pm p(y) \wedge \pm r(x,y)) \wedge \pm q(x))$.
        \end{tabbing}
    \end{minipage}
\end{center}
\noindent 

The sentence templates for $\mathcal{A}$ are those of $\mathcal{Z}$ together with
\begin{center}
\begin{minipage}{8cm}
\begin{tabbing}
Some/\= every (non-) $o$ $r$'s some/every (non-)\\ 
\> $p$ who $s$'s/does not $s$ him\\
No (non-) $o$ $r$'s any/every (non-) $p$ who \\
\> $s$'s/does not $s$ him\\
Some (non-) $o$ does not $r$ any/every (non-) $p$ \\
\> who $s$'s/does not $s$ him.
\end{tabbing} 
\end{minipage}
\end{center}
We can define the first-order formulae for giving semantics for the above sentence templates as follows:
\begin{center}
    \begin{minipage}{7cm}
        \begin{tabbing}
            $\forall x(\pm o(x) \rightarrow \forall y(\pm p(y) \wedge \pm s(y,x) \rightarrow \pm r(x,y)))$ \\
            $\forall x(\pm o(x) \rightarrow \exists y(\pm p(y) \wedge \pm s(y,x) \wedge \pm r(x,y)))$ \\
            $\exists x(\pm o(x) \wedge \forall y(\pm p(y) \wedge \pm s(y,x) \rightarrow \pm r(x,y)))$ \\
            $\exists x(\pm o(x) \wedge \forall y(\pm p(y) \wedge \pm s(y,x) \wedge \pm r(x,y)))$.
        \end{tabbing}
    \end{minipage}
\end{center}


\section{Appendix: Computational Complexity of $Sat(\mathcal{S})$, $Sat(\mathcal{W})$, $Sat(\mathcal{V})$, $Sat(\mathcal{Z})$ and $Sat(\mathcal{A})$}
\label{sec:appendix_complexity}

In this section, we justify the complexity-theoretic information provided in section \ref{sec:languageFragments}. The case of $\mathcal{S}$
requires no work: this is the fragment denoted $\mathcal{S}^\dagger$ in~\cite{pratt2009logics}, where the complexity of satisfiability is established in Theorem~4.10, p.~668. For $\mathcal{W}$, the upper bound of \NPTime{} follows easily from the fact that all the relevant formulae are in the 1-variable fragment of first-order logic; the lower bound follows by a straightforward reduction from 3-SAT, as given, for example, for the
smaller fragment denoted $\mathcal{E}_1$ in~\cite[Theorem 2, p.~213]{pratt2004fragments}. The fragment $\mathcal{V}$ is slightly larger than the fragment denoted $\mathcal{E}_2$ in~\cite{pratt2004fragments}, whose satisfiability problem is shown to be \ExpTime{}-complete in Theorem~3, p.215.
This settles the lower bound for Sat($\mathcal{V}$). For the upper bound, we note that the proof depends essentially on the fact that all of the first-order formulae involved contain just one occurrence of any binary predicate; therefore, the same proof applies to the fragment
$\mathcal{V}$ almost without change of wording. Furthermore, the same argumentation applies to $\mathcal{Z}$. The only remaining case to consider
is $\mathcal{A}$, which is dealt with by the following theorem.

\textbf{Theorem 1} The computational complexity of the problem of determining satisfiability in a set of sentences in $\mathcal{A}$ is \textsc{NExpTime} complete

\textbf{Proof: } A moment's thought shows that every sentence of $\mathcal{A}$ is expressed by a formula of the two-variable fragment of first-order
logic, $\mathcal{FO}^2$, whose satisfiability problem is \NExpTime-complete~\cite{gkv97}. 
Thus, $\Sat(\mathcal{A})$ is in \NExpTime.
To establish a matching lower bound
we take an existing proof of the \NExpTime-hardness of $\Sat(\mathcal{FO}^2)$, and show that it can be reproduced using.
only the resources of $\mathcal{A}$. The proof in question proceeds by reduction from a known \NExpTime-hard
problem $\mathcal{P}$ 
(in this case, $\mathcal{P}$ is the problem of determining whether an exponential-sized grid can be tiled
in a certain tiling system), and encoding any instance of{} $\mathcal{P}$ as a set $\Phi$ of $\mathcal{FO}^2$-formulae such that $\Phi$ is satisfiable if and only if the given instance of $\mathcal{P}$ is positive. It can be
shown that such encodings (see for example~\citealt[pp.~90~ff.]{pratt2023FOL}) require only that we can write 
$\mathcal{FO}^2$-formulae of the 
following forms:
\begin{align}
\exists x (p_1(x) \wedge \cdots \wedge \pm p_n(x))
\label{eq:form1}\\
\forall x (\pm p_1(x) \vee \cdots \vee \pm p_n(x))
\label{eq:form2}\\
\forall x \forall y (\bigwedge_{i=1}^n(p_i(x) \leftrightarrow q_i(y)) \rightarrow r(x,y))
\label{eq:form3}\\
\forall x (\pm p(x) \rightarrow \exists y (\pm q(y) \wedge \pm r(x,y)))
\label{eq:form4}\\
\forall x (\pm p(x) \rightarrow \forall y (\pm q(y) \rightarrow \pm r(x,y))).
\label{eq:form5}
\end{align}
Thus, if we can reproduce the effect of such formulae in $\mathcal{A}$, then we will have established that
$\Sat(\mathcal{A})$ is \NExpTime-hard.

Form~\eqref{eq:form1} can be simulated by a collection of sentences of the forms
``{Some $p^*$ is a $p^*$}'',
``{Every $p^*$ is a $p_1$}'', \dots ``{Every $p^*$ is a $p_n$}'', where $p^*$ is a fresh unary predicate/common noun. Form~\eqref{eq:form2} is handled similarly, remembering that noun-level negation is available. Any formula of Form~\eqref{eq:form3} can be simulated by
the formulae $\forall x \forall y (r_n(x,y) \rightarrow r(x,y))$ and
$\forall x \forall y\, r_0(x,y)$ together with the collection of formulae 
$\forall x(\pm_1 p_i(x) \rightarrow  \forall y (\pm_1 q_i(y) \wedge r_{i-1}(x,y) \rightarrow r_i(x,y))$ ($1 \leq i \leq  n$),
where the two occurrences of $\pm_1$ are resolved in the same way, and the $r_0, \dots, r_n$ are new binary predicates/transitive verbs.
Of these, the formula $\forall x \forall y\, r_0(x,y)$ is equivalent to the conjunction of the four formulae
$\forall x (\pm q(x) \rightarrow \forall y (\pm q(x) \rightarrow r_0(x,y))$, which
can be written directly in $\mathcal{A}$, while the remaining formulae  can be written easily in $\mathcal{A}$.
Forms~\eqref{eq:form4} and~\eqref{eq:form5} can be written directly in $\mathcal{A}$.

Thus, any instance of an exponential tiling problem can be transformed, in polynomial time, to an instance of Sat($\mathcal{A}$) with the same answer.
This establishes the \NExpTime-hardness of Sat($\mathcal{A}$).


\section{Appendix: Zero-shot Prompting Templates}
\label{sec:zeroshot_templates}

Given a set of sentences ${s_1, s_2, ..., s_m}$ and a label $\ell$ indicating whether the set of sentences is satisfiable or not, we formulate the prompt using the following template. 

Q: Given the following set of sentences, tell me whether they are satisfiable or not. Generate satisfiable if they are and unsatisfiable if they are not. 
Set of sentences: ${s_1, s_2, ..., s_m}$

A:

As we mentioned in the Results and Discussion section, the LLaMa model did not generate ``satisfiable'' or ``unsatisfiable'' for the above prompt for most problem instances. Therefore, we 
modified
the prompt by following an example problem instance and asking it to generate the words ``True'' or ``False'' instead. 
The example problem instance consists of a set of sentences ${\overbar{s}_1, \overbar{s}_2, ..., \overbar{s}_m}$ and a satisfiability label be $\overbar{\ell}$ (where $\overbar{\ell}$ is ``True'' if ${\overbar{s}_1, \overbar{s}_2, ..., \overbar{s}_m}$ is satisfiable and ``False'' otherwise). 
The resultant prompt is as follows:
	
Q: Given the following set of sentences, tell me whether they are satisfiable or not. Generate True if they are and False if they are not. 

Set of sentences: ${\overbar{s}_1, \overbar{s}_2, ..., \overbar{s}_m}$

A: $\overbar{\ell}$

Given the following set of sentences, tell me whether they are satisfiable or not. Generate True if they are and False if they are not. 

Set of sentences: ${s_1, s_2, ..., s_m}$

A: 

\section{Appendix: Dataset and Training Details}
\label{sec:appendix_training}

\subsection{Dataset details}

We utilised five different language fragments, namely, Syllogistic $\mathcal{S}$, relative clauses $\mathcal{W}$, relative clauses with relational syllogistic $\mathcal{V}$, relative clauses with transitive verbs $\mathcal{Z}$ and anaphora $\mathcal{A}$. For each language fragment, we constructed a training set with 120K, an eval set with 10K and a test set with 10K data points. When constructing the dataset, we set the number of unary predicates $n_1$ to be between $3$ and $8$ and the number of binary predicates $n_2$ to be between $3$ and $8$ for fragments $\mathcal{V}$, $\mathcal{Z}$ and $\mathcal{A}$ while setting the $n_1$ value to be between 6 and 16 for fragments $\mathcal{S}$ and $\mathcal{W}$. It is apparent that the distribution of the number of ways specific numbers between $6$ and $16$ can be formed by randomly adding two numbers in the range $3$ and $8$ formed a bell curve. Hence, the distribution of the number of problem instances per number of variables of $\mathcal{V}$, $\mathcal{Z}$ and $\mathcal{A}$ seems to be a discrete approximation of normal distribution. Therefore, we sample the $n_1$ values from a normal distribution for $\mathcal{S}$ and $\mathcal{W}$ while sampling $n_1$, $n_2$ values uniformly for the other fragments. 

When testing for TLMs' ability to generalise, we construct a separate test set with more predicates than that of the training set. We construct 6K data points for each fragment and the dataset is balanced having an equal number of satisfiable and unsatisfiable problem instances. In this setup, we set the range of the $n_1$ value in the test set between $16$ and $24$ for fragments $\mathcal{S}$, and $\mathcal{W}$, while varying it between $8$ and $12$ for other fragments. The range of the number of binary predicates $n_2$ for fragments $\mathcal{V}$, $\mathcal{Z}$ and $\mathcal{A}$ was set to be between $8$ and $24$ due to the asymmetric nature of the variation of probability of satisfiability with $\frac{m}{n_1}$ compared to $\frac{m}{n_2}$. It is important to emphasise that all data points belonging to each of the datasets (train, eval and test) are unique. 

To evaluate TLMs in zero-shot settings, we construct a separate test set with 1200 data points for each fragment, with 600 data points being satisfiable while the other data points being unsatisfiable. In this setup, we set $n_1^{min}$ = 5, and $n_1^{max}$ = 10 for fragments $\mathcal{S}$ and $\mathcal{W}$ and $n_1^{min}$ = 3, $n_1^{max}$ = 5, $n_2^{min}$ = 2 and $n_2^{max}$ = 5 for fragments $\mathcal{V}$, $\mathcal{Z}$, and $\mathcal{A}$. . Moreover, for each value $n$ where $n = n_1 + n_2$, the dataset contains 200 data points, with 100 of them being satisfiable and the other 100 being unsatisfiable. 

When defining the vocabulary, we expanded the vocabulary introduced by Richardson and Sabharwal \shortcite{richardson2021pushing}. The vocabulary of unary predicates $\mathcal{U}$ comprises 156 nouns (of people's professions) and the vocabulary of binary predicates $\mathcal{V}$ comprises 70 transitive verbs. 

The mean, maximum, and minimum number of clauses for each of the language fragments are depicted in Table \ref{table:m_values}, while mean, maximum, and minimum instance lengths are illustrated in Table \ref{table:num_tokens}. 

\begin{table}
\centering
  \begin{tabular}{|m{1.5cm}|m{1.5cm}|m{1.5cm}|m{1.5cm}|} 
  \hline
  \makecell{\textbf{Language} \\ \textbf{Fragment}} & \makecell{\textbf{minimum}} & \makecell{\textbf{maximum}} & \makecell{\textbf{mean}} \\ \hline

\makecell{$\mathcal{S}$} & \makecell{29} & \makecell{8} & \makecell{17.06}\\ \hline
\makecell{$\mathcal{W}$} & \makecell{32} & \makecell{9} & \makecell{18.95} \\ \hline
\makecell{$\mathcal{V}$} &\makecell{25} & \makecell{5} & \makecell{14.00}\\ \hline
\makecell{$\mathcal{Z}$} & \makecell{28} & \makecell{6} & \makecell{14.31}\\ \hline
\makecell{$\mathcal{A}$} & \makecell{34} & \makecell{8} & \makecell{16.50}\\ \hline

  \end{tabular}
  \newline
  \caption{minimum, maximum and mean number of clauses $m$ in the training set for the fragments we consider, $\mathcal{S}$, $\mathcal{W}$, $\mathcal{V}$, $\mathcal{Z}$ and $\mathcal{A}$}
  \label{table:m_values}
\end{table}

\begin{table}
\centering
  \begin{tabular}{|m{1.5cm}|m{1.5cm}|m{1.5cm}|m{1.5cm}|} 
  \hline
  \makecell{\textbf{Language} \\ \textbf{Fragment}} & \makecell{\textbf{minimum}} & \makecell{\textbf{maximum}} & \makecell{\textbf{mean}} \\ \hline

\makecell{$\mathcal{S}$} & \makecell{149} & \makecell{35} & \makecell{83.19}\\ \hline
\makecell{$\mathcal{W}$} & \makecell{267} & \makecell{43} & \makecell{129.08} \\ \hline
\makecell{$\mathcal{V}$} &\makecell{221} & \makecell{22} & \makecell{97.37}\\ \hline
\makecell{$\mathcal{Z}$} & \makecell{274} & \makecell{29} & \makecell{110.22}\\ \hline
\makecell{$\mathcal{A}$} & \makecell{318} & \makecell{39} & \makecell{123.42}\\ \hline

  \end{tabular}
  \newline
  \caption{minimum, maximum and mean number of words (tokens) when separated by \texttt{SPACE} in the training set for the language fragments}
  \label{table:num_tokens}
\end{table}

\begin{figure*}
    \centering
    \includegraphics[width = 11.5cm]{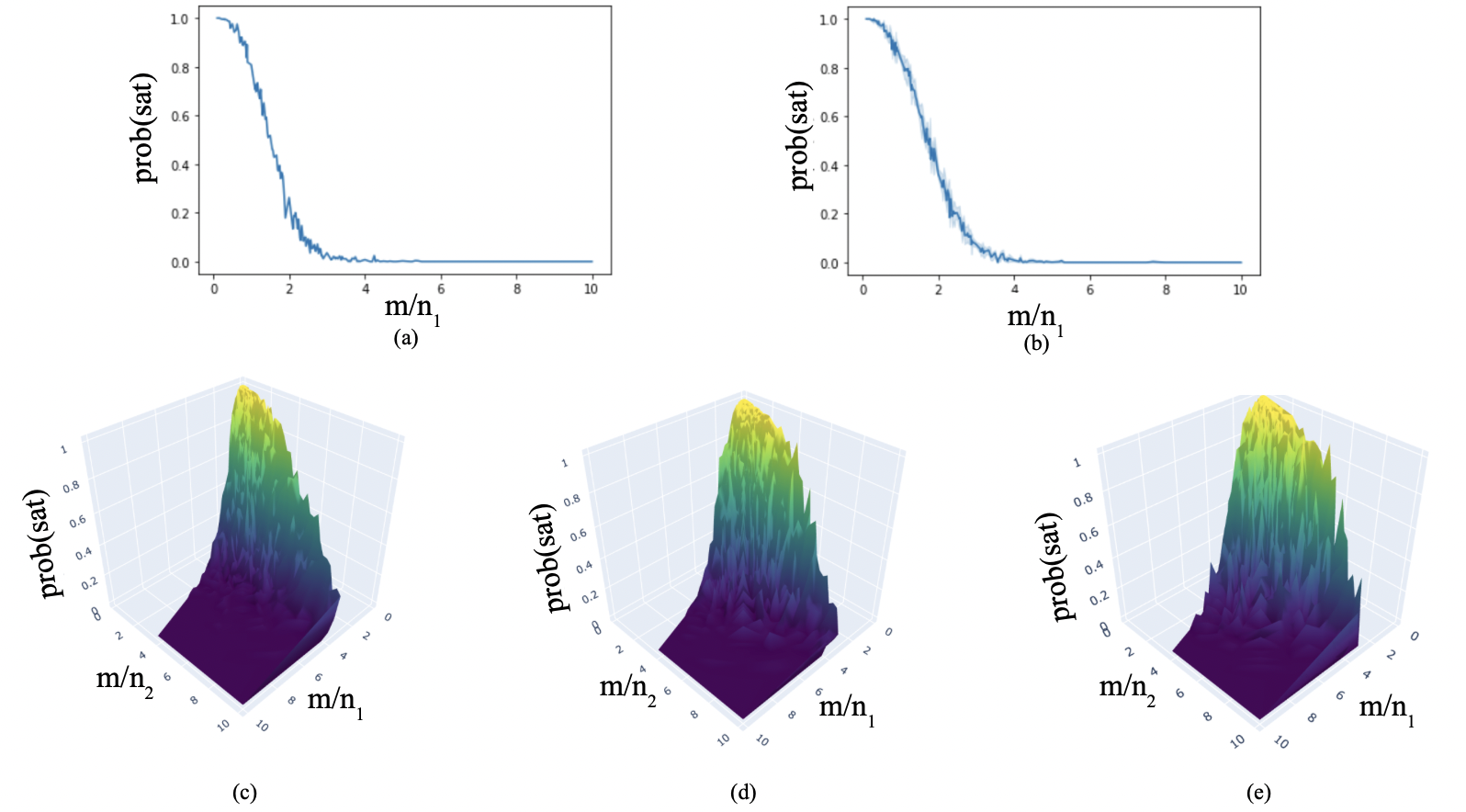}
    \caption{The variation of probability of satisfiability with clause unary and binary variable ratios for the language fragments (a) $\mathcal{S}$, (b) $\mathcal{W}$, (c) $\mathcal{V}$, (d) $\mathcal{Z}$ and (e) $\mathcal{A}$. The symbol $m$ denotes the number of clauses, and $a$, $b$ indicate the number of unary and binary variables respectively. }
    \label{fig:prob_sat_ab}
\end{figure*}

\subsection{Training Details}

We chose two TLMs with somewhat different architectures; T5 and DeBERTa-v3, as both models have proven track records on textual entailment tasks. Since the research question of interest relies on the learnability of TLMs, not on identifying the best-performing model, we do not perform any hyperparameter tuning. Since according to prior literature \cite{raffel2019exploring}, the performance of the TLMs stems from model size and pre-trained data more so than the architectural choice, as well as, since the accuracy values yielded for the TLMs are similar, we anticipated similar performance for other TLMs. Moreover, exploring different transformer models and hyper-parameter tuning would leave a higher carbon footprint \cite{strubell-etal-2019-energy}, which is deemed unnecessary considering the nature of the research questions. 

A detailed description of the hyperparameters is as follows,

\paragraph{Maximum sequence length :} When training the DeBERTa-v3 model, we used the maximum sequence length between 512 and 700 tokens, and when training the T5 model using the joint dataset, we used 700 tokens as the maximum sequence length. Since we used the T5 models trained on individual fragments for testing for generalisation, we used a maximum sequence length of 1024 tokens. We did not rely on any truncation, as truncating input could alter the satisfiability of the input sentences. 
\paragraph{Training epochs :} We used five training epochs and used the eval dataset to identify the best-performing model.
\paragraph{Batch size :} Relying on the gradient checkpointing, we used a batch size of 24 for the DeBERTa-v3-large model. Similarly, when trained with the joint dataset, we used a batch size of 24, again relying on gradient checkpointing. Since we used the T5 model trained on the individual dataset for generalisation experiments, we downgraded the batch size to 12 to prevent memory errors.  

Each of the TLM is fined-tuned to predict the label by reducing the binary cross entropy loss over the target using Adam optimiser \cite{DBLP:journals/corr/KingmaB14} and we used the HuggingFace \cite{wolf2019huggingface} implementation when experimenting with the above-mentioned TLM.

\section{Appendix: Phase Change Region of Language Fragments}
\label{sec:appendix_phase_change_region}

\begin{figure*}
    \centering
    \includegraphics[width = 13.5cm]{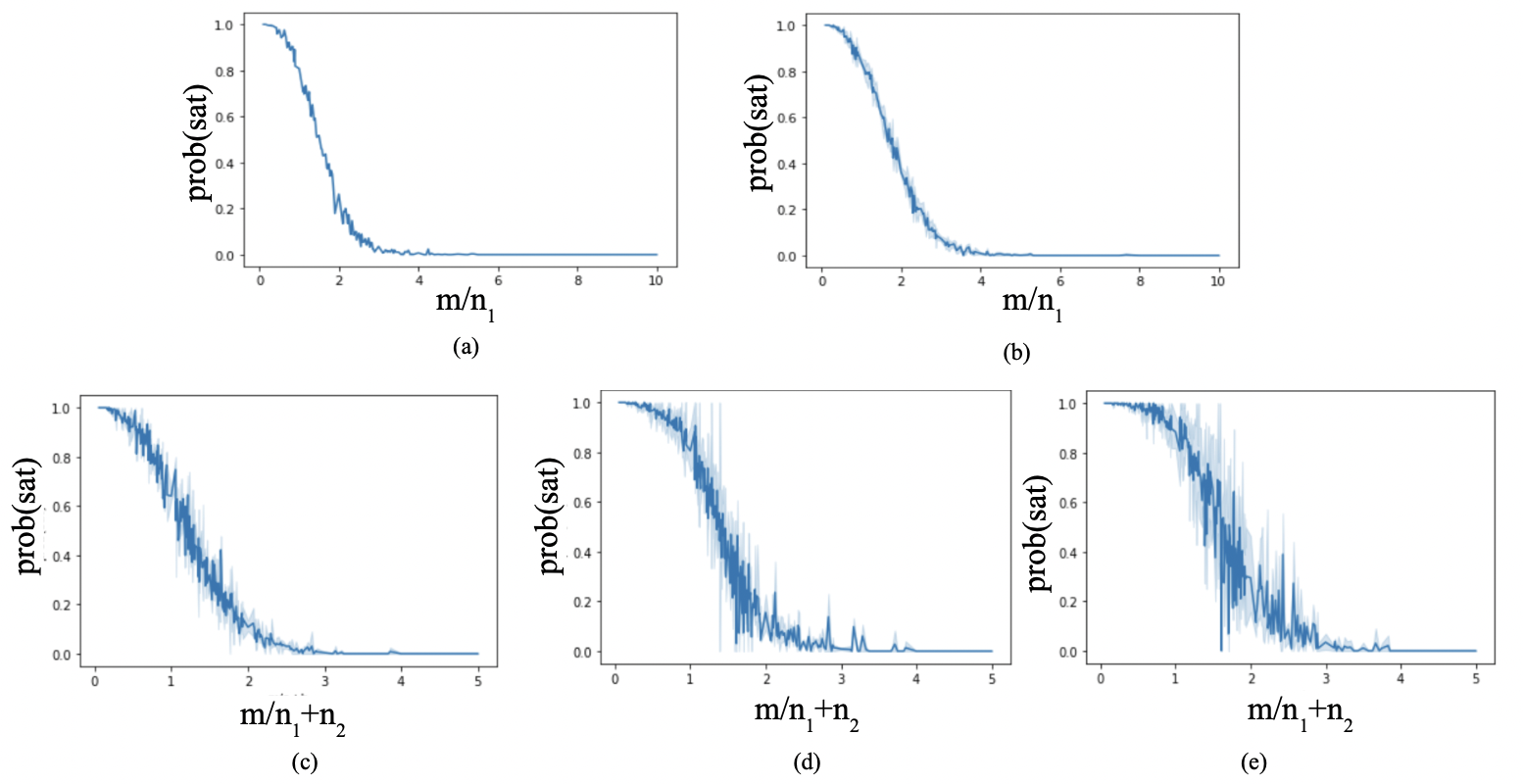}
    \caption{The variation of probability of satisfiability with clause variable ratio for the language fragments (a) $\mathcal{S}$, (b) $\mathcal{W}$, (c) $\mathcal{V}$, (d) $\mathcal{Z}$ and (e) $\mathcal{A}$, where we consider the total variables $a+b$.}
    \label{fig:prob_sat_m_n}
\end{figure*}

\begin{figure*}
    \centering
    \includegraphics[width = 11.5cm]{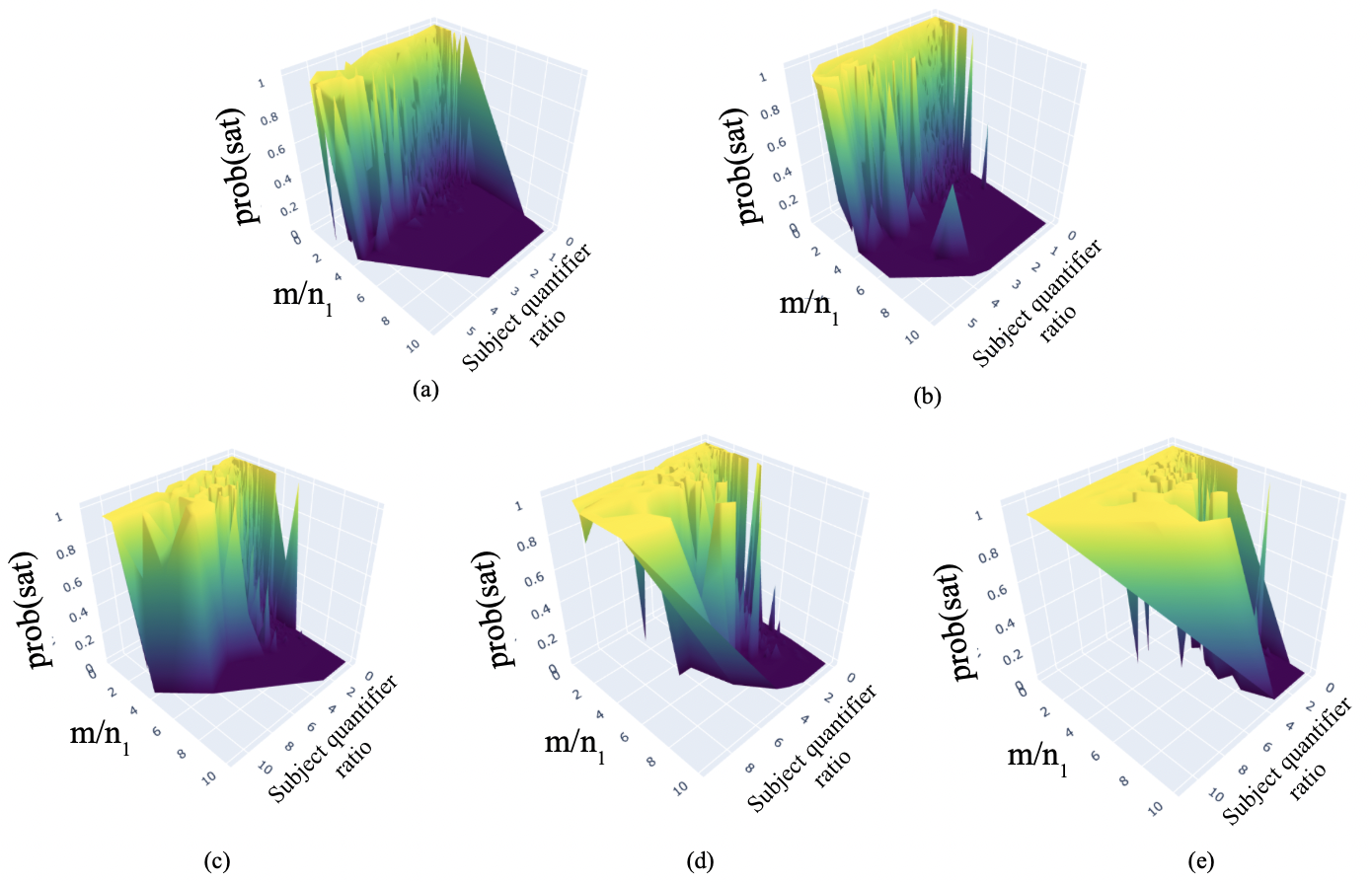}
    \caption{The variation of probability of satisfiability with clause unary variable ratio and subject quantifier ration for the language fragments (a) $\mathcal{S}$, (b) $\mathcal{W}$, (c) $\mathcal{V}$, (d) $\mathcal{Z}$ and (e) $\mathcal{A}$. The symbol $m$ denotes the number of clauses, and $a$ indicates the number of unary variables respectively. }
    \label{fig:prob_sat_ma_sub}
\end{figure*}

\begin{figure*}
    \centering
    \includegraphics[width = 11.5cm]{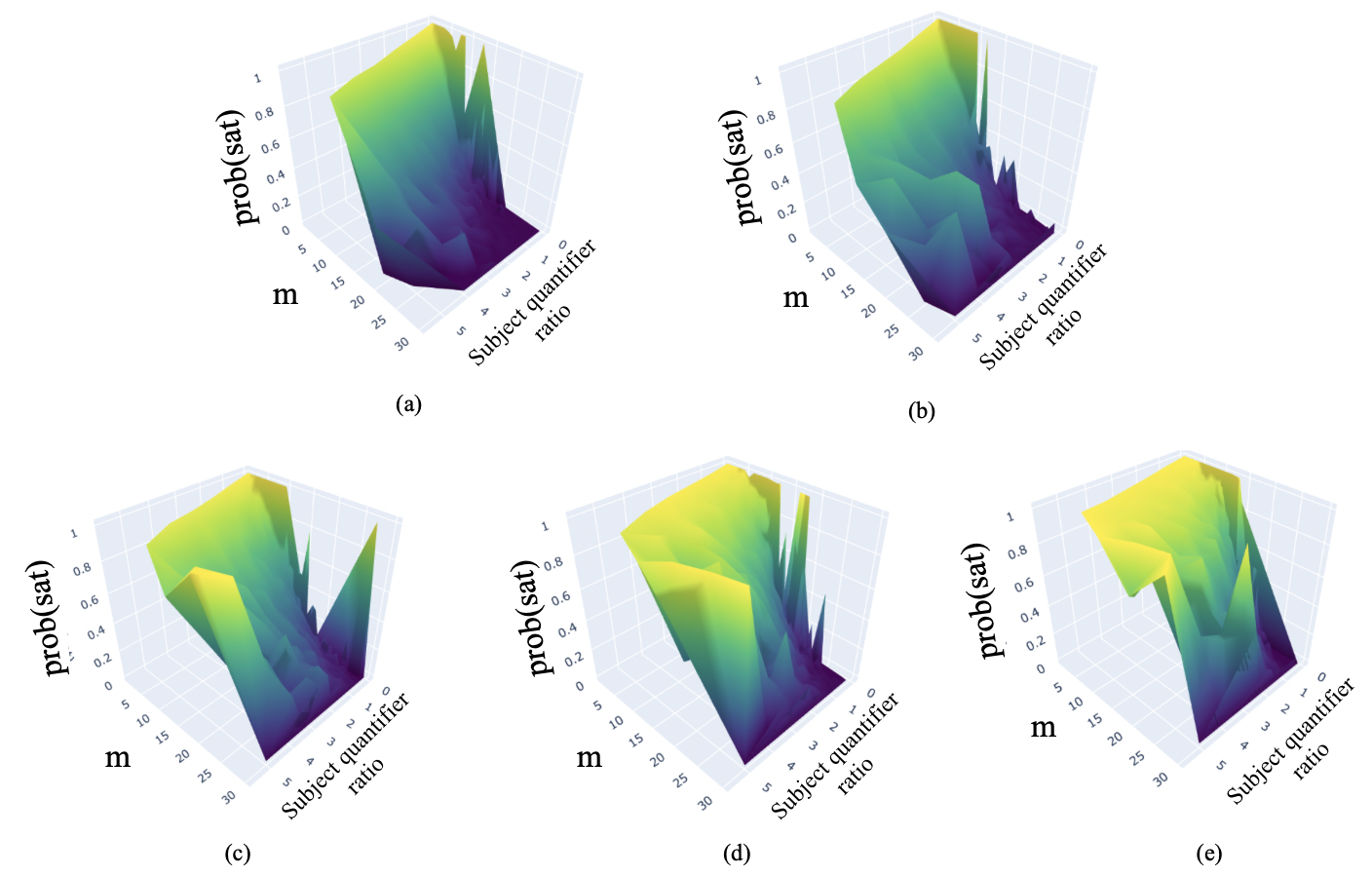}
    \caption{The variation of probability of satisfiability with the number of clauses and subject quantifier ratio for the language fragments (a) $\mathcal{S}$, (b) $\mathcal{W}$, (c) $\mathcal{V}$, (d) $\mathcal{Z}$ and (e) $\mathcal{A}$. The symbol $m$ denotes the number of clauses.}
    \label{fig:prob_m_sub}
\end{figure*}

\begin{figure*}
    \centering
    \includegraphics[width = 11.5cm]{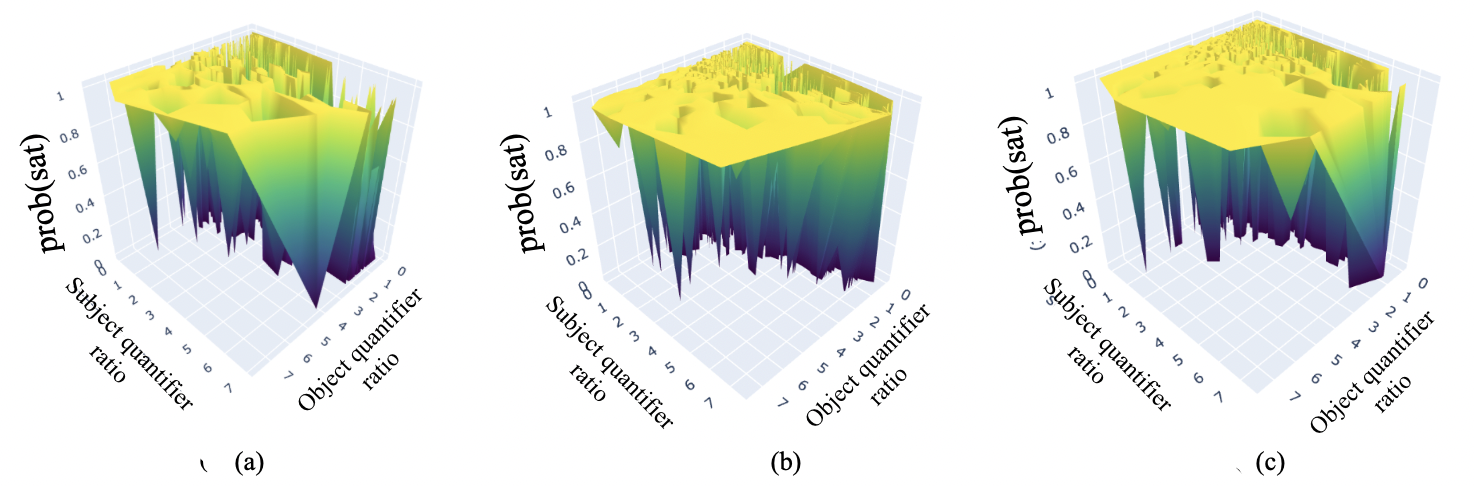}
    \caption{The variation of probability of satisfiability with subject quantifier ratio and object quantifier ratio for the language fragments (a) $\mathcal{V}$, (b) $\mathcal{Z}$ and (c) $\mathcal{A}$}
    \label{fig:prob_sat_sub_obj}
\end{figure*}

We explore the problem distribution of the language fragments we considered along several analytical viewpoints. The quantifier ratio is used to define the ratio of the number of $\exists$ quantifiers to the number of $\forall$ quantifiers, and the subject quantifier ratio is the quantifier ratio between the quantifiers associated with the subject while the object quantifier ratio is that associated with the object. Since sentences belonging to the fragments $\mathcal{S}$ and $\mathcal{W}$ contain only one quantifier, which is associated with the subject, the object quantifier ratio is omitted for those fragments. The visualisation of variation in the probability of satisfiability is depicted in figures \ref{fig:prob_sat_ab}, \ref{fig:prob_sat_m_n}, \ref{fig:prob_sat_ma_sub}, \ref{fig:prob_m_sub} and \ref{fig:prob_sat_sub_obj}. Figure \ref{fig:prob_sat_ab} illustrates how the probability of satisfiability varies with $\frac{m}{n_1}$ and $\frac{m}{n_2}$, which we have used when defining the phase change region.

\begin{thebibliography}{50}
\expandafter\ifx\csname natexlab\endcsname\relax\def\natexlab#1{#1}\fi

\bibitem[{Aristotle(1938)}]{aristotle}
Aristotle. 1938.
\newblock \emph{The Categories, On Interpretation, Prior Analytics}.
\newblock Loeb Classical Library. Harvard University Press, Cambridge, MA.
\newblock Tr.~H.~Cooke and H.~Tredennick.

\bibitem[{Bang et~al.(2023)Bang, Cahyawijaya, Lee, Dai, Su, Wilie, Lovenia, Ji, Yu, Chung et~al.}]{bang2023multitask}
Yejin Bang, Samuel Cahyawijaya, Nayeon Lee, Wenliang Dai, Dan Su, Bryan Wilie, Holy Lovenia, Ziwei Ji, Tiezheng Yu, Willy Chung, et~al. 2023.
\newblock A multitask, multilingual, multimodal evaluation of chatgpt on reasoning, hallucination, and interactivity.
\newblock \emph{arXiv preprint arXiv:2302.04023}.

\bibitem[{Beltagy et~al.(2020)Beltagy, Peters, and Cohan}]{Beltagy2020Longformer}
Iz~Beltagy, Matthew~E Peters, and Arman Cohan. 2020.
\newblock Longformer: The long-document transformer.
\newblock \emph{arXiv preprint arXiv:2004.05150}.

\bibitem[{Cameron et~al.(2020{\natexlab{a}})Cameron, Chen, Hartford, and Leyton-Brown}]{cameron2020predicting}
Chris Cameron, Rex Chen, Jason Hartford, and Kevin Leyton-Brown. 2020{\natexlab{a}}.
\newblock Predicting propositional satisfiability via end-to-end learning.
\newblock In \emph{Proceedings of the AAAI Conference on Artificial Intelligence}, volume~34, pages 3324--3331.

\bibitem[{Cameron et~al.(2020{\natexlab{b}})Cameron, Chen, Hartford, and Leyton-Brown}]{Cameron2020PredictingPS}
Chris Cameron, Rex Chen, Jason~S. Hartford, and Kevin Leyton-Brown. 2020{\natexlab{b}}.
\newblock Predicting propositional satisfiability via end-to-end learning.
\newblock In \emph{AAAI}.

\bibitem[{Chen et~al.(2020)Chen, Li, and Sun}]{chen2020differentiable}
Ting Chen, Lala Li, and Yizhou Sun. 2020.
\newblock Differentiable product quantization for end-to-end embedding compression.
\newblock In \emph{International Conference on Machine Learning}, pages 1617--1626. PMLR.

\bibitem[{Clark et~al.(2021)Clark, Tafjord, and Richardson}]{clark2020transformers}
Peter Clark, Oyvind Tafjord, and Kyle Richardson. 2021.
\newblock Transformers as soft reasoners over language.
\newblock In \emph{Proceedings of the Twenty-Ninth International Joint Conference on Artificial Intelligence}, IJCAI'20.

\bibitem[{Cook and Mitchell(1997)}]{cook1997finding}
Stephen~A Cook and David~G Mitchell. 1997.
\newblock Finding hard instances of the satisfiability problem: A survey.
\newblock \emph{Satisfiability Problem: Theory and Applications}, 35:1--17.

\bibitem[{Devlin et~al.(2018)Devlin, Chang, Lee, and Toutanova}]{devlin2018bert}
Jacob Devlin, Ming-Wei Chang, Kenton Lee, and Kristina Toutanova. 2018.
\newblock Bert: Pre-training of deep bidirectional transformers for language understanding.
\newblock \emph{arXiv preprint arXiv:1810.04805}.

\bibitem[{Evans et~al.(2018)Evans, Saxton, Amos, Kohli, and Grefenstette}]{evans2018can}
Richard Evans, David Saxton, David Amos, Pushmeet Kohli, and Edward Grefenstette. 2018.
\newblock \href {https://openreview.net/forum?id=SkZxCk-0Z} {Can neural networks understand logical entailment?}
\newblock In \emph{International Conference on Learning Representations}.

\bibitem[{Geiger et~al.(2018)Geiger, Cases, Karttunen, and Potts}]{geiger2018stress}
Atticus Geiger, Ignacio Cases, Lauri Karttunen, and Christopher Potts. 2018.
\newblock Stress-testing neural models of natural language inference with multiply-quantified sentences.
\newblock \emph{arXiv preprint arXiv:1810.13033}.

\bibitem[{Glockner et~al.(2018)Glockner, Shwartz, and Goldberg}]{glockner-etal-2018-breaking}
Max Glockner, Vered Shwartz, and Yoav Goldberg. 2018.
\newblock \href {https://doi.org/10.18653/v1/P18-2103} {Breaking {NLI} systems with sentences that require simple lexical inferences}.
\newblock In \emph{Proceedings of the 56th Annual Meeting of the Association for Computational Linguistics (Volume 2: Short Papers)}, pages 650--655, Melbourne, Australia. Association for Computational Linguistics.

\bibitem[{Goodwin et~al.(2020)Goodwin, Sinha, and O{'}Donnell}]{goodwin-etal-2020-probing}
Emily Goodwin, Koustuv Sinha, and Timothy~J. O{'}Donnell. 2020.
\newblock \href {https://doi.org/10.18653/v1/2020.acl-main.177} {Probing linguistic systematicity}.
\newblock In \emph{Proceedings of the 58th Annual Meeting of the Association for Computational Linguistics}, pages 1958--1969, Online. Association for Computational Linguistics.

\bibitem[{Gr{\"a}del et~al.(1997)Gr{\"a}del, Kolaitis, and Vardi}]{gkv97}
E.~Gr{\"a}del, P.~Kolaitis, and M.~Vardi. 1997.
\newblock On the decision problem for two-variable first-order logic.
\newblock \emph{Bulletin of Symbolic Logic}, 3(1):53--69.

\bibitem[{He and Choi(2020)}]{he-choi-flairs-2020}
Han He and Jinho~D. Choi. 2020.
\newblock \href {https://aaai.org/ocs/index.php/FLAIRS/FLAIRS20/paper/view/18438} {{Establishing Strong Baselines for the New Decade: Sequence Tagging, Syntactic and Semantic Parsing with BERT}}.
\newblock In \emph{Proceedings of the 33rd International Florida Artificial Intelligence Research Society Conference}, FLAIRS'20, pages 228--233.

\bibitem[{He et~al.(2021)He, Gao, and Chen}]{he2021debertav3}
Pengcheng He, Jianfeng Gao, and Weizhu Chen. 2021.
\newblock Debertav3: Improving deberta using electra-style pre-training with gradient-disentangled embedding sharing.
\newblock \emph{arXiv preprint arXiv:2111.09543}.

\bibitem[{Kamath and Das(2019)}]{kamath2019a}
Aishwarya Kamath and Rajarshi Das. 2019.
\newblock \href {https://openreview.net/forum?id=HylaEWcTT7} {A survey on semantic parsing}.
\newblock In \emph{Automated Knowledge Base Construction (AKBC)}.

\bibitem[{Kingma and Ba(2015)}]{DBLP:journals/corr/KingmaB14}
Diederik~P. Kingma and Jimmy Ba. 2015.
\newblock \href {http://arxiv.org/abs/1412.6980} {Adam: {A} method for stochastic optimization}.
\newblock In \emph{3rd International Conference on Learning Representations, {ICLR} 2015, San Diego, CA, USA, May 7-9, 2015, Conference Track Proceedings}.

\bibitem[{Lake and Baroni(2017)}]{Lake2017GeneralizationWS}
Brenden~M. Lake and Marco Baroni. 2017.
\newblock Generalization without systematicity: On the compositional skills of sequence-to-sequence recurrent networks.
\newblock In \emph{International Conference on Machine Learning}.

\bibitem[{Lin et~al.(2019)Lin, Tafjord, Clark, and Gardner}]{lin-etal-2019-reasoning}
Kevin Lin, Oyvind Tafjord, Peter Clark, and Matt Gardner. 2019.
\newblock \href {https://doi.org/10.18653/v1/D19-5808} {Reasoning over paragraph effects in situations}.
\newblock In \emph{Proceedings of the 2nd Workshop on Machine Reading for Question Answering}, pages 58--62, Hong Kong, China. Association for Computational Linguistics.

\bibitem[{Liu et~al.(2019)Liu, Ott, Goyal, Du, Joshi, Chen, Levy, Lewis, Zettlemoyer, and Stoyanov}]{liu2019roberta}
Yinhan Liu, Myle Ott, Naman Goyal, Jingfei Du, Mandar Joshi, Danqi Chen, Omer Levy, Mike Lewis, Luke Zettlemoyer, and Veselin Stoyanov. 2019.
\newblock Roberta: A robustly optimized bert pretraining approach.
\newblock \emph{arXiv preprint arXiv:1907.11692}.

\bibitem[{Madusanka et~al.(2023{\natexlab{a}})Madusanka, Batista-navarro, and Pratt-hartmann}]{madusanka-etal-2023-identifying}
Tharindu Madusanka, Riza Batista-navarro, and Ian Pratt-hartmann. 2023{\natexlab{a}}.
\newblock \href {https://doi.org/10.18653/v1/2023.eacl-main.257} {Identifying the limits of transformers when performing model-checking with natural language}.
\newblock In \emph{Proceedings of the 17th Conference of the European Chapter of the Association for Computational Linguistics}, pages 3539--3550, Dubrovnik, Croatia. Association for Computational Linguistics.

\bibitem[{Madusanka et~al.(2023{\natexlab{b}})Madusanka, Zahid, Li, Pratt-Hartmann, and Batista-Navarro}]{madusanka-etal-2023-quantifiers}
Tharindu Madusanka, Iqra Zahid, Hao Li, Ian Pratt-Hartmann, and Riza Batista-Navarro. 2023{\natexlab{b}}.
\newblock \href {https://doi.org/10.18653/v1/2023.emnlp-main.536} {Not all quantifiers are equal: Probing transformer-based language models{'} understanding of generalised quantifiers}.
\newblock In \emph{Proceedings of the 2023 Conference on Empirical Methods in Natural Language Processing}, pages 8680--8692, Singapore. Association for Computational Linguistics.

\bibitem[{McCoy et~al.(2019)McCoy, Pavlick, and Linzen}]{mccoy-etal-2019-right}
Tom McCoy, Ellie Pavlick, and Tal Linzen. 2019.
\newblock \href {https://doi.org/10.18653/v1/P19-1334} {Right for the wrong reasons: Diagnosing syntactic heuristics in natural language inference}.
\newblock In \emph{Proceedings of the 57th Annual Meeting of the Association for Computational Linguistics}, pages 3428--3448, Florence, Italy. Association for Computational Linguistics.

\bibitem[{Minervini et~al.(2020)Minervini, Bosnjak, Rockt{\"{a}}schel, Riedel, and Grefenstette}]{DBLP:conf/aaai/MinerviniBR0G20}
Pasquale Minervini, Matko Bosnjak, Tim Rockt{\"{a}}schel, Sebastian Riedel, and Edward Grefenstette. 2020.
\newblock \href {https://ojs.aaai.org/index.php/AAAI/article/view/5962} {Differentiable reasoning on large knowledge bases and natural language}.
\newblock In \emph{The Thirty-Fourth {AAAI} Conference on Artificial Intelligence, {AAAI} February 7-12, 2020}, pages 5182--5190. {AAAI} Press.

\bibitem[{Mitchell and Levesque(1996)}]{mitchell1996some}
David~G Mitchell and Hector~J Levesque. 1996.
\newblock Some pitfalls for experimenters with random sat.
\newblock \emph{Artificial Intelligence}, 81(1-2):111--125.

\bibitem[{Narodytska et~al.(2020)Narodytska, Zhang, Gupta, and Walsh}]{Narodytska2020In}
Nina Narodytska, Hongce Zhang, Aarti Gupta, and Toby Walsh. 2020.
\newblock \href {https://openreview.net/forum?id=SJx-j64FDr} {In search for a sat-friendly binarized neural network architecture}.
\newblock In \emph{International Conference on Learning Representations}.

\bibitem[{OpenAI(2023)}]{openai2023gpt4}
OpenAI. 2023.
\newblock \href {http://arxiv.org/abs/2303.08774} {Gpt-4 technical report}.

\bibitem[{Pratt-Hartmann(2003)}]{pratt2003two}
Ian Pratt-Hartmann. 2003.
\newblock A two-variable fragment of english.
\newblock \emph{Journal of Logic, Language and Information}, 12(1):13--45.

\bibitem[{Pratt-Hartmann(2004)}]{pratt2004fragments}
Ian Pratt-Hartmann. 2004.
\newblock Fragments of language.
\newblock \emph{Journal of Logic, Language and Information}, 13(2):207--223.

\bibitem[{Pratt-Hartmann(2023)}]{pratt2023FOL}
Ian Pratt-Hartmann. 2023.
\newblock \emph{Fragments of First-Order Logic}.
\newblock Oxford University Press, Oxford, UK.

\bibitem[{Pratt-Hartmann and Moss(2009)}]{pratt2009logics}
Ian Pratt-Hartmann and Lawrence~S Moss. 2009.
\newblock Logics for the relational syllogistic.
\newblock \emph{The Review of Symbolic Logic}, 2(4):647--683.

\bibitem[{Pratt-Hartmann and Third(2006)}]{pratt2006more}
Ian Pratt-Hartmann and Allan Third. 2006.
\newblock More fragments of language.
\newblock \emph{Notre Dame Journal of Formal Logic}, 47(2):151--177.

\bibitem[{Raffel et~al.(2019)Raffel, Shazeer, Roberts, Lee, Narang, Matena, Zhou, Li, and Liu}]{raffel2019exploring}
Colin Raffel, Noam Shazeer, Adam Roberts, Katherine Lee, Sharan Narang, Michael Matena, Yanqi Zhou, Wei Li, and Peter~J Liu. 2019.
\newblock Exploring the limits of transfer learning with a unified text-to-text transformer.
\newblock \emph{arXiv preprint arXiv:1910.10683}.

\bibitem[{Richardson et~al.(2020)Richardson, Hu, Moss, and Sabharwal}]{richardson2020probing}
Kyle Richardson, Hai Hu, Lawrence Moss, and Ashish Sabharwal. 2020.
\newblock Probing natural language inference models through semantic fragments.
\newblock In \emph{Proceedings of the AAAI Conference on Artificial Intelligence}, volume~34, pages 8713--8721.

\bibitem[{Richardson and Sabharwal(2021)}]{richardson2021pushing}
Kyle Richardson and Ashish Sabharwal. 2021.
\newblock Pushing the limits of rule reasoning in transformers through natural language satisfiability.
\newblock \emph{arXiv preprint arXiv:2112.09054}.

\bibitem[{Saha et~al.(2020)Saha, Ghosh, Srivastava, and Bansal}]{saha-etal-2020-prover}
Swarnadeep Saha, Sayan Ghosh, Shashank Srivastava, and Mohit Bansal. 2020.
\newblock \href {https://doi.org/10.18653/v1/2020.emnlp-main.9} {{PR}over: Proof generation for interpretable reasoning over rules}.
\newblock In \emph{Proceedings of the 2020 Conference on Empirical Methods in Natural Language Processing (EMNLP)}, pages 122--136, Online. Association for Computational Linguistics.

\bibitem[{Selman et~al.(1996)Selman, Mitchell, and Levesque}]{selman1996generating}
Bart Selman, David~G Mitchell, and Hector~J Levesque. 1996.
\newblock Generating hard satisfiability problems.
\newblock \emph{Artificial intelligence}, 81(1-2):17--29.

\bibitem[{Selsam et~al.(2019)Selsam, Lamm, B\"{u}nz, Liang, de~Moura, and Dill}]{selsam2018learning}
Daniel Selsam, Matthew Lamm, Benedikt B\"{u}nz, Percy Liang, Leonardo de~Moura, and David~L. Dill. 2019.
\newblock \href {https://openreview.net/forum?id=HJMC_iA5tm} {Learning a {SAT} solver from single-bit supervision}.
\newblock In \emph{International Conference on Learning Representations}.

\bibitem[{Shin et~al.(2019)Shin, Kant, Gupta, Bender, Trabucco, Singh, and Song}]{shin2018synthetic}
Richard Shin, Neel Kant, Kavi Gupta, Chris Bender, Brandon Trabucco, Rishabh Singh, and Dawn Song. 2019.
\newblock \href {https://openreview.net/forum?id=ryeOSnAqYm} {Synthetic datasets for neural program synthesis}.
\newblock In \emph{International Conference on Learning Representations}.

\bibitem[{Strubell et~al.(2019)Strubell, Ganesh, and McCallum}]{strubell-etal-2019-energy}
Emma Strubell, Ananya Ganesh, and Andrew McCallum. 2019.
\newblock \href {https://doi.org/10.18653/v1/P19-1355} {Energy and policy considerations for deep learning in {NLP}}.
\newblock In \emph{Proceedings of the 57th Annual Meeting of the Association for Computational Linguistics}, pages 3645--3650, Florence, Italy. Association for Computational Linguistics.

\bibitem[{Tafjord et~al.(2021)Tafjord, Dalvi, and Clark}]{Tafjord2021ProofWriterGI}
Oyvind Tafjord, Bhavana Dalvi, and Peter Clark. 2021.
\newblock \href {https://doi.org/10.18653/v1/2021.findings-acl.317} {{P}roof{W}riter: Generating implications, proofs, and abductive statements over natural language}.
\newblock In \emph{Findings of the Association for Computational Linguistics: ACL-IJCNLP 2021}, pages 3621--3634, Online. Association for Computational Linguistics.

\bibitem[{Touvron et~al.(2023)Touvron, Martin, Stone, Albert, Almahairi, Babaei, Bashlykov, Batra, Bhargava, Bhosale et~al.}]{touvron2023llama}
Hugo Touvron, Louis Martin, Kevin Stone, Peter Albert, Amjad Almahairi, Yasmine Babaei, Nikolay Bashlykov, Soumya Batra, Prajjwal Bhargava, Shruti Bhosale, et~al. 2023.
\newblock Llama 2: Open foundation and fine-tuned chat models.
\newblock \emph{arXiv preprint arXiv:2307.09288}.

\bibitem[{Weber et~al.(2019)Weber, Minervini, M{\"u}nchmeyer, Leser, and Rockt{\"a}schel}]{weber-etal-2019-nlprolog}
Leon Weber, Pasquale Minervini, Jannes M{\"u}nchmeyer, Ulf Leser, and Tim Rockt{\"a}schel. 2019.
\newblock \href {https://doi.org/10.18653/v1/P19-1618} {{NLP}rolog: Reasoning with weak unification for question answering in natural language}.
\newblock In \emph{Proceedings of the 57th Annual Meeting of the Association for Computational Linguistics}, pages 6151--6161, Florence, Italy. Association for Computational Linguistics.

\bibitem[{Welleck et~al.(2021)Welleck, Liu, Bras, Hajishirzi, Choi, and Cho}]{welleck2021naturalproofs}
Sean Welleck, Jiacheng Liu, Ronan~Le Bras, Hannaneh Hajishirzi, Yejin Choi, and Kyunghyun Cho. 2021.
\newblock \href {https://openreview.net/forum?id=Jvxa8adr3iY} {Naturalproofs: Mathematical theorem proving in natural language}.
\newblock In \emph{Thirty-fifth Conference on Neural Information Processing Systems Datasets and Benchmarks Track (Round 1)}.

\bibitem[{Wolf et~al.(2019)Wolf, Debut, Sanh, Chaumond, Delangue, Moi, Cistac, Rault, Louf, Funtowicz et~al.}]{wolf2019huggingface}
Thomas Wolf, Lysandre Debut, Victor Sanh, Julien Chaumond, Clement Delangue, Anthony Moi, Pierric Cistac, Tim Rault, R{\'e}mi Louf, Morgan Funtowicz, et~al. 2019.
\newblock Huggingface's transformers: State-of-the-art natural language processing.
\newblock \emph{arXiv preprint arXiv:1910.03771}.

\bibitem[{Wu et~al.(2021)Wu, Kreiss, Ong, and Potts}]{wu2021reascan}
Zhengxuan Wu, Elisa Kreiss, Desmond Ong, and Christopher Potts. 2021.
\newblock \href {https://openreview.net/forum?id=Rtquf4Jk0jN} {Rea{SCAN}: Compositional reasoning in language grounding}.
\newblock In \emph{Thirty-fifth Conference on Neural Information Processing Systems Datasets and Benchmarks Track (Round 1)}.

\bibitem[{Xu et~al.(2020)Xu, Li, Zhang, Du, ichi Kawarabayashi, and Jegelka}]{Xu2020What}
Keyulu Xu, Jingling Li, Mozhi Zhang, Simon~S. Du, Ken ichi Kawarabayashi, and Stefanie Jegelka. 2020.
\newblock \href {https://openreview.net/forum?id=rJxbJeHFPS} {What can neural networks reason about?}
\newblock In \emph{International Conference on Learning Representations}.

\bibitem[{Yanaka et~al.(2020)Yanaka, Mineshima, Bekki, and Inui}]{yanaka-etal-2020-neural}
Hitomi Yanaka, Koji Mineshima, Daisuke Bekki, and Kentaro Inui. 2020.
\newblock \href {https://doi.org/10.18653/v1/2020.acl-main.543} {Do neural models learn systematicity of monotonicity inference in natural language?}
\newblock In \emph{Proceedings of the 58th Annual Meeting of the Association for Computational Linguistics}, pages 6105--6117, Online. Association for Computational Linguistics.

\bibitem[{Yang et~al.(2019)Yang, Dai, Yang, Carbonell, Salakhutdinov, and Le}]{yang2019xlnet}
Zhilin Yang, Zihang Dai, Yiming Yang, Jaime Carbonell, Russ~R Salakhutdinov, and Quoc~V Le. 2019.
\newblock Xlnet: Generalized autoregressive pretraining for language understanding.
\newblock \emph{Advances in neural information processing systems}, 32.

\end{thebibliography}
\end{document}